\documentclass[journal]{IEEEtran}

\usepackage{amsmath,amsfonts}
\usepackage{algorithmic}
\usepackage{array}
\usepackage[caption=false,font=normalsize,labelfont=sf,textfont=sf]{subfig}
\usepackage{textcomp}
\usepackage{stfloats}
\usepackage{url}
\usepackage{verbatim}
\usepackage{graphicx}
\usepackage{cite}
\usepackage{amsmath,amssymb,amsfonts}
\usepackage{algorithm}
\usepackage{algorithmic}
\usepackage[hyperfootnotes=false]{hyperref}
\usepackage{graphicx}
\usepackage{textcomp}
\usepackage{xcolor}
\usepackage{amsmath}
\usepackage{cleveref}
\usepackage{multirow}
\usepackage{enumitem}
\usepackage{setspace}
\usepackage{url}
\usepackage{booktabs}
\usepackage{tabularx}

\definecolor{lightred}{rgb}{1, 0.3, 0.3}
\definecolor{darkblue}{rgb}{0, 0, 0.5}
\newcommand{\best}[1]{\textcolor{lightred}{\textbf{#1}}}

\def\BibTeX{{\rm B\kern-.05em{\sc i\kern-.025em b}\kern-.08em
    T\kern-.1667em\lower.7ex\hbox{E}\kern-.125emX}}
\usepackage{balance}
\begin{document}
\title{Efficient and Scalable Multiview Masked Autoencoder for Heterogeneous Graph Self-Supervised Learning}
\author{Hao Geng, Deqing Wang, Lihao Ge, Chenguang Du, Fuzhen Zhuang,~\IEEEmembership{Senior Member,~IEEE,}\\ Bowen Yang, Yanlin Hu, and Songtao Wang\thanks{Received 23 October 2024; revised 18 July 2025 and 23 February 2026; accepted 13 October 2026. This work was supported in part by the National Natural Science Foundation of China under Grant 62276015 and in part by the Fundamental Research Funds for the Central Universities under Grant GW2025-09. (Corresponding author: Songtao Wang.)}\thanks{Hao Geng and Lihao Ge are with Beihang University, Beijing 100191, China (e-mail: genghao@buaa.edu.cn; glh6660815@gmail.com).}\thanks{Deqing Wang and Fuzhen Zhuang are with Beihang University, Beijing 100191, China, and also with the Zhongguancun Laboratory, Beijing 100194, China (e-mail: dqwang@buaa.edu.cn; zhuangfuzhen@buaa.edu.cn).}\thanks{Chenguang Du and Songtao Wang are with the Zhongguancun Laboratory, Beijing 100194, China (e-mail: duchenguang@buaa.edu.cn; wangst@zgclab.edu.cn).}\thanks{Bowen Yang is with the Westlake University, Hangzhou 310030, China (e-mail: by296@cantab.ac.uk).}\thanks{Yanlin Hu is with the National Computer Network Emergency Response Technical Team/Coordination Center of China, Beijing 100029, China (e-mail: yanlinhu@cert.org.cn).}\thanks{Digital Object Identifier 10.1109/TNNLS.2026.3745261.}}

\markboth{IEEE Transactions on Neural Networks and Learning Systems}{Geng \MakeLowercase{\textit{et al.}}: Efficient and Scalable Multiview Masked Autoencoder}

\hypersetup{hidelinks,bookmarksdepth=2,
  pdftitle={Efficient and Scalable Multiview Masked Autoencoder for Heterogeneous Graph Self-Supervised Learning},
  pdfauthor={Hao Geng; Deqing Wang; Lihao Ge; Chenguang Du; Fuzhen Zhuang; Bowen Yang; Yanlin Hu; Songtao Wang}}

\maketitle

\begin{abstract}
Recent years have witnessed the flourishing of self-supervised learning (SSL) on heterogeneous graphs, e.g., masked autoencoder (MAE)-based methods. However, existing MAE-based methods mostly suffer from three issues, including the prohibitive computational cost of constructing meta-path-induced subgraphs, the limitations of uniform random masking, and the suboptimal objective of reconstructing potentially noisy or uninformative original node features.
To address these issues, we propose the multiview masked autoencoder (MVMAE)---a novel, efficient, and scalable framework for heterogeneous graph SSL.
First, MVMAE introduces a novel multiview mask-then-reconstruct strategy. Instead of masking individual nodes or features, MVMAE randomly masks entire meta-path views and guides the model to reconstruct their representations from remaining visible views. This approach shifts the learning objective to a high-level, semantic task, providing a more robust and informative supervisory signal that is effective even when initial node features are absent or flawed.
Then, to handle the computational demands, MVMAE employs an efficient offline aggregation mechanism that precomputes and reuses representations for each meta-path view, completely avoiding the costly construction of meta-path-induced subgraphs.
Next, to fuse the representations of each view, MVMAE uses a Transformer-based multiview fusion encoder to capture the complex correlations between views, and a deliberately simple MLP decoder to compel the encoder to learn highly expressive and robust node representations.
Extensive experiments on {six} real-world datasets, including {three} large-scale graphs, demonstrate that MVMAE consistently outperforms state-of-the-art self-supervised and even semisupervised baselines, confirming its superior effectiveness and scalability.
Our source code is released at \url{https://github.com/hcmdgh/MVMAE}.
\end{abstract}


\section{Introduction} \label{sec:intro}

Recent years have witnessed the flourishing of heterogeneous graph representation learning (HGRL) in data mining and knowledge discovery \cite{DBLP:journals/tbd/WangBSFYY23,DBLP:conf/aaai/YangYPYF23}.
A heterogeneous graph is a potent data structure for representing diverse types of entities and complex relationships between them.
Consequently, numerous real-world systems can be modeled as heterogeneous graphs, enabling a wide range of applications in domains such as recommender systems \cite{DBLP:conf/wsdm/ChenHXWXL23}, bioinformatics \cite{DBLP:journals/bib/YiYHK22}, and academic networks \cite{DBLP:journals/tkde/ZhangLTDYZGWKSLW23}.
For effectively modeling heterogeneous graphs, heterogeneous graph neural networks (HGNNs) have emerged as a state-of-the-art methodology, demonstrating remarkable efficacy in capturing rich structural and semantic information \cite{DBLP:conf/www/WangJSWYCY19,DBLP:conf/www/HuDWS20,DBLP:conf/aaai/YangYPYF23}.
However, most HGNNs operate under supervised or semisupervised paradigms \cite{DBLP:conf/aaai/0001DZZC23}, where the learning process relies heavily on labeled data. This reliance presents a significant practical bottleneck, as label acquisition in real-world applications is often challenging and prohibitively expensive \cite{DBLP:journals/tbd/WangBSFYY23}.
To mitigate this issue, self-supervised learning (SSL) offers a promising solution by learning supervisory signals directly from the data itself, bypassing the need for explicit labels.

Contrastive learning methods dominate graph SSL \cite{DBLP:conf/www/0001XYLWW21,DBLP:conf/icml/YouCSW21,DBLP:journals/corr/abs-2405-11868}, but their success depends on challenging data augmentation and negative sampling strategies \cite{DBLP:conf/kdd/HouLCDYW022}.
Most graph augmentations, like node dropping or edge perturbation, are heuristic-based, leading to inconsistent performance. This makes finding an optimal strategy a costly trial-and-error process with no guarantee of generalizability \cite{DBLP:conf/icml/YouCSW21,DBLP:journals/sigkdd/DingXTL22}. Furthermore, negative sampling is often computationally demanding, and suboptimal sampling strategies can introduce uninformative samples that degrade model performance \cite{DBLP:conf/kdd/WangLHS21}.

Generative SSL methods circumvent the issues associated with contrastive approaches by learning representations through the direct reconstruction of input data \cite{DBLP:conf/kdd/HouLCDYW022, DBLP:conf/www/HouHCLDK023,DBLP:conf/icml/VincentLBM08,DBLP:journals/corr/KipfW16a}. Among these, the masked autoencoder (MAE) framework has proven particularly effective, initially demonstrating remarkable success in computer vision \cite{DBLP:conf/cvpr/HeCXLDG22} and natural language processing \cite{DBLP:conf/naacl/DevlinCLT19} by learning to predict masked portions of the input data.
Inspired by these advancements, the MAE paradigm is adapted to the graph domain. For instance, GraphMAE \cite{DBLP:conf/kdd/HouLCDYW022} learns to reconstruct randomly masked node features.
HGMAE \cite{DBLP:conf/aaai/0001DZZC23} further extends the approach to heterogeneous graphs by reconstructing both masked nodes and edges in each meta-path-induced subgraph.
However, despite their promising performance on certain graph learning tasks, existing generative SSL methods overlook critical challenges inherent in real-world graphs, thereby limiting their practical applicability.

\begin{figure}[tp]
    \centering
    \includegraphics[width=1\columnwidth]{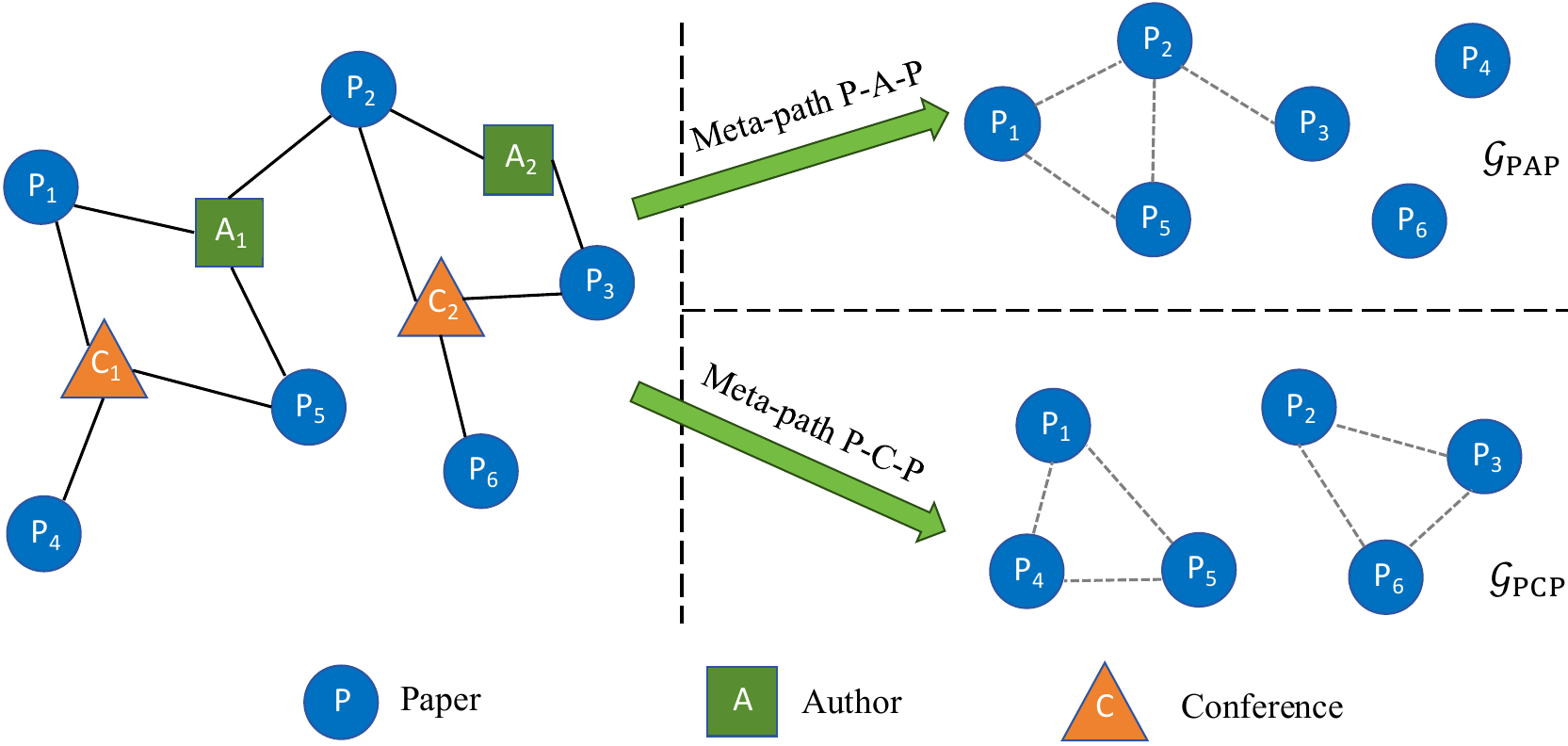}
        \caption{
    Example of meta-path-induced subgraphs in OAG.
    $\mathcal{G}_{\text{PAP}}$ and $\mathcal{G}_{\text{PCP}}$ are two subgraphs induced by meta-paths P-A-P and P-C-P, respectively.
    }
    \label{fig:subgraph}
\end{figure}

{\textbf{C1: Storage and Computational Bottlenecks via Metapath-induced Density Explosion:}
A meta-path is a sequence of relations defining a composite relation between node types, serving as a high-level blueprint for capturing complex semantics in heterogeneous graphs. To model graph heterogeneity, most existing graph representation learning methods decompose the heterogeneous graph into multiple meta-path-induced subgraphs. However, this process often encounters severe scalability issues in practice.
As illustrated in Fig.~\ref{fig:subgraph}, in datasets like the Open Academic Graph (OAG) \cite{DBLP:journals/tkde/ZhangLTDYZGWKSLW23}, meta-paths such as paper-conference-paper (P-C-P) can lead to a ``density explosion'' of the resulting adjacency matrices. Due to the ``small-world effect'' and the prevalence of high-degree nodes (e.g., a single conference or author linked to thousands of papers), the number of meta-path instances representing the nonzero (nnz) elements in the induced subgraph can grow even quadratically. For instance, a meta-path connecting 10,000 papers via a shared conference can generate $10^8$ edges, producing a dense intermediate matrix that far exceeds the memory capacity of modern GPUs.
Consequently, while sparse matrix multiplication (SpMM) theoretically avoids explicit $O(|\mathcal{V}|^2)$ construction, the prohibitive storage and computational overhead of these dense meta-path-induced edges severely hinder the scalability of existing methods to large-scale heterogeneous graphs.}

{\textbf{C2: Limitations of Microlevel Uniform Random Masking in Graph SSL:}
Existing graph SSL methods, including GraphMAE \cite{DBLP:conf/kdd/HouLCDYW022} and HGMAE \cite{DBLP:conf/aaai/0001DZZC23}, typically rely on a uniform random masking strategy at the microlevel. This technique involves randomly masking a portion of individual node features or edges and training the model to reconstruct the masked content.
While straightforward, a critical flaw of this strategy lies in its exclusive focus on isolated, microlevel components. By restricting the masking target to individual nodes or edges, these methods often fail to challenge the model to capture complex, high-level structural semantics. For instance, within a densely interconnected community of a graph where node features tend to be highly homogeneous, predicting a masked node's feature from its local neighbors is a relatively trivial task.
Such a scenario creates an overly simplistic reconstruction objective, which severely constrains the model's ability to learn expressive, robust, and discriminative representations. To overcome this limitation, it is crucial to shift the masking target from microlevel elements to macrolevel structures (e.g., meta-path views) and introduce challenging cross-semantic reasoning into the reconstruction process.}

\textbf{C3: Limitations of Reconstructing Original Node Features in Graph SSL:}
Another significant issue arises from the common practice of reconstructing original node features as the self-supervised objective in graph SSL methods. While intuitive, this paradigm has two primary weaknesses.
First, the original node features themselves may be inherently flawed. These features can often contain noise, irrelevant information, or be incomplete.
For example, in the AMiner dataset, all nodes lack initial features. Researchers often use a random Gaussian distribution to generate them, with no semantic meaning.
When the model is tasked with reconstructing these corrupted or deficient features, the learning process can be misguided, leading to poor performance in downstream tasks.
Second, original node features exclusively describe the intrinsic properties of individual nodes.
This narrow focus neglects the broader contextual information derived from a node's position and relationships within the graph's topology.
By tasking the model solely with the reconstruction of these intrinsic attributes, crucial information is overlooked, including the complex correlations between interconnected nodes and the rich structural and semantic information embedded in the graph structure.
Consequently, the model's capacity to learn a comprehensive and robust representation of the graph is severely constrained.

To address these issues, we propose the multiview masked autoencoder (MVMAE) model for heterogeneous graph SSL. We treat each meta-path as a distinct ``view'' of the heterogeneous graph, as each meta-path view encapsulates a specific structural aspect of the graph.

To address C1, we propose an efficient offline aggregation mechanism that completely avoids the explicit construction of costly meta-path-induced subgraphs. Inspired by the simplification of graph convolutions \cite{DBLP:conf/icml/WuSZFYW19,DBLP:journals/corr/abs-2004-11198}, our approach computes aggregated features for each meta-path view via a sequence of sparse-dense matrix multiplications.
This process has a linear complexity in the number of edges. It involves no trainable parameters, allowing it to be performed as a one-time offline precomputation. This drastically reduces the computational burden during training while maintaining remarkable performance.

To address C2 and C3, we introduce a multiview mask-then-reconstruct strategy. Instead of masking individual nodes or edges, we randomly mask a portion of the meta-path views. The model is then tasked with reconstructing the representations of these masked views using the remaining visible views.
This approach shifts the learning objective from a low-level and uniform masking task to a high-level and semantic one.
Furthermore, rather than reconstructing flawed original features (as pointed out in C3), the model learns to reconstruct the semantically rich meta-path view representations generated by our offline aggregator. This provides a stable and informative supervisory signal, enabling robust learning even on graphs that lack meaningful initial features.
The intuition behind our strategy is that the semantics of each meta-path are often correlated. For example, in an academic graph, this paper-author-paper (P-A-P) meta-path and the paper-paper (P-P) meta-path both contain semantic signals that can help to predict this paper-field-paper (P-F-P) meta-path, as authors tend to publish in and cite papers from the same field.
By explicitly training the model to find these connections, it learns a deeper understanding of the graph's structure.

To fuse multiple meta-path views into a unified representation, we propose a multiview fusion encoder that leverages the Transformer architecture~\cite{DBLP:conf/nips/VaswaniSPUJGKP17}.
For each node, we treat its set of view embeddings as a sequence of tokens. Then a self-attention mechanism captures the correlations between these views, fusing them into a single and unified node representation.
Next, the fused representation is fed to a simple MLP decoder to perform the reconstruction task. The deliberate simplicity of the MLP decoder forces the encoder to learn highly expressive and informative representations, as it cannot rely on a powerful decoder to compensate for information loss.

Our main contributions are summarized as follows:
\begin{enumerate}\item We propose a novel self-supervised method for heterogeneous graphs---masking and reconstructing meta-path views. Since meta-path views contain richer structural and semantic information, this strategy enables the model to capture rich semantic and structural information in the whole graph.
\item To avoid the prohibitive cost of constructing meta-path-induced subgraphs, we design an efficient offline aggregation mechanism that can be precomputed with a complexity linear in the number of edges.
\item We propose to utilize multiview representations generated by a parameter-free mechanism as reconstruction targets. This not only provides an informative and robust supervision signal but also elegantly addresses the ``trivial solution'' problem in the MAE paradigm.
\item Extensive experiments on five real-world datasets demonstrate that our proposed model not only outperforms state-of-the-art unsupervised baselines but also surpasses several supervised counterparts. Furthermore, its strong performance on two large-scale datasets confirms its scalability and effectiveness.
\end{enumerate}

\section{Related Work}

In this section, we review existing work related to our study, specifically focusing on HGNNs and SSL on graphs.

\subsection{Heterogeneous Graph Neural Networks}
In recent years, the increasing prevalence of data structured as heterogeneous graphs has led to the emergence of HGNNs \cite{DBLP:journals/tnn/WuPCLZY21,DBLP:journals/tkde/ZhangCZ22,DBLP:journals/tist/KhoshraftarA24,DBLP:conf/kdd/LvDLCFHZJDT21,DBLP:journals/tkde/YangGLZCW23}. These models aim to learn powerful node representations by capturing complex structural and semantic information.
The majority of existing work in this area has primarily focused on supervised learning frameworks.
For example, the heterogeneous graph attention network (HAN) \cite{DBLP:conf/www/WangJSWYCY19} utilizes meta-path-based attention to differentiate node and semantic importance.
MAGNN \cite{DBLP:conf/www/0004ZMK20} enriches meta-path representations by aggregating features from intermediate nodes.
HGT \cite{DBLP:conf/www/HuDWS20} proposed node-level and relation-level attention mechanisms based on Transformer \cite{DBLP:conf/nips/VaswaniSPUJGKP17}, which can be applied to large-scale heterogeneous graphs.
{SHGN \cite{DBLP:conf/kdd/LvDLCFHZJDT21} attempts to enhance efficiency by converting heterogeneous graphs into homogeneous ones with learnable edge-type embeddings. However, it bypasses the explicit modeling of high-order semantic information provided by meta-paths.}

A common flaw among these methods is their dependence on labeled data for training. This supervised paradigm limits their applicability in scenarios where labels are scarce or unavailable and prevents them from fully exploiting the intrinsic attributes and topological features of the graph.

\subsection{SSL on Graphs}
The substantial labor and financial costs associated with manual data annotation have spurred a recent surge in SSL, which leverages a dataset's intrinsic properties for model training \cite{DBLP:journals/tkde/LiuZHMWZT23,DBLP:journals/pami/GuiCZCSLT24,DBLP:conf/cvpr/HeCXLDG22}.
Within the domain of graph data, SSL methods are broadly categorized into two paradigms: contrastive and generative learning.

Graph contrastive learning operates by sampling positive and negative pairs from the graph and then applying an objective function to minimize the distance between positive samples while maximizing the distance between negative ones \cite{DBLP:conf/www/JingPT21,DBLP:conf/iclr/VelickovicFHLBH19,DBLP:journals/corr/abs-1911-08538,DBLP:conf/aaai/ParkK0Y20,DBLP:conf/kdd/WangLHS21}.
A foundational approach in this area is DGI \cite{DBLP:conf/iclr/VelickovicFHLBH19}, which maximizes the mutual information between local patch representations and a global summary of the entire graph.
HDGI \cite{DBLP:journals/corr/abs-1911-08538} adapts the DGI framework for heterogeneous graphs. It utilizes meta-paths to model complex semantics and employs a semantic-level attention mechanism to learn node representations.
DMGI \cite{DBLP:conf/aaai/ParkK0Y20} introduces a consensus regularization framework and a universal discriminator to jointly integrate embeddings from different relation types.
HeCo \cite{DBLP:conf/kdd/WangLHS21} also employs a contrastive framework, utilizing two views to align node representations.
Despite their success, a primary challenge for these contrastive methods is their heavy reliance on the quality of negative samples. Effective negative sampling strategies often vary across different graph structures, posing an obstacle to the development of universally applicable models.

Alternatively, generative learning paradigms primarily employ an autoencoder framework \cite{DBLP:conf/cvpr/HeCXLDG22}. It involves an encoder that maps the input graph to latent representations and a decoder tasked with reconstructing masked or corrupted portions of the original data from these representations.
One prominent line of work focuses on feature reconstruction.
For instance, GraphMAE \cite{DBLP:conf/kdd/HouLCDYW022} utilizes an MAE to reconstruct original node features and achieves strong performance. However, its performance can be vulnerable to the discriminability of the input features.
In contrast to feature-centric methods, MaskGAE \cite{DBLP:conf/kdd/LiWS0TZMZW23} pioneers a generative approach focused on structure reconstruction. It introduces Masked Graph Modeling (MGM), a pretext task where a portion of the graph's edges are masked and then reconstructed.
When applying these generative models to heterogeneous graphs, existing approaches like HGMAE \cite{DBLP:conf/aaai/0001DZZC23} have proposed multiple tasks, such as meta-path-based edge and attribute restoration.
Nevertheless, these tasks often operate at different granularities, which can impede model convergence. Furthermore, such methods typically perform masking within individual meta-paths without explicitly modeling the crucial associations between them.
{More recently, Hetero2Net \cite{li2025heterophily} introduces a masked meta-path prediction task to learn local topologies via subgraph sampling.
However, a critical limitation of such sampling-based generative methods is the inevitable information loss. Due to the "small-world effect," restricting the number of sampled neighbors to maintain efficiency often leads to the omission of vital structural and semantic context. Furthermore, existing methods often fail to explicitly model the synergistic associations between different meta-path views.}

Building on these insights, in this work we introduce a novel generative learning task and framework designed to overcome the limitations of prior methods.

\section{Preliminaries}

In this section, we first introduce the concepts of heterogeneous graphs and meta-path. Then we provide a formal definition for the problem of generative SSL on heterogeneous graphs.

\subsection{Heterogeneous Graph} A heterogeneous graph is defined as a graph $\mathcal{G} = (\mathcal{V}, \mathcal{E}, \mathcal{A}, \mathcal{T}_\mathcal{V}, \mathcal{T}_\mathcal{E}, \mathbf{X})$. It consists of multiple types of nodes represented by $\mathcal{V}$ and edges represented by $\mathcal{E}$, and an adjacency matrix $\mathcal{A}$ that captures the relationship between nodes. The sets $\mathcal{T}_\mathcal{V}$ and $\mathcal{T}_\mathcal{E}$ include the types of nodes and edges, respectively. In addition, the node feature matrix set $\mathbf{X}= \{ \mathbf{X}_{\mathcal{T}_{i}} \in \mathbb{R}^{|\mathcal{V}_{\mathcal{T}_{i}}| \times d_{\mathcal{T}_{i}}} \}$ denotes attributes associated with each node. Here, $\mathcal{T}_{i}$ refers to a specific node type, and $d_{\mathcal{T}_{i}}$ represents the dimension of the corresponding node attributes.

\subsection{Meta-Path}
In a heterogeneous graph $\mathcal{G}$, the schema is defined by a set of node types $\mathcal{T}_\mathcal{A}$ and a set of relation types $\mathcal{T}_\mathcal{E}$. The fundamental building block of this schema is the meta-relation, which describes a directed relation between a source node type and a target node type.
Formally, for an edge $e=(u, v)$ of type $\rho(e) \in \mathcal{T}_\mathcal{E}$ that connects a source node $u$ of type $\tau(u) \in \mathcal{T}_\mathcal{A}$ to a target node $v$ of type $\tau(v) \in \mathcal{T}_\mathcal{A}$, its meta-relation is represented by the triplet $(\tau(u), \rho(e), \tau(v))$.
Building upon this, a meta-path $\phi$ is defined as a sequence of meta-relations that specifies a composite path between node types. A meta-path of length $l$ is denoted as $\phi = A_1 \xrightarrow{R_1} A_2 \xrightarrow{R_2} \cdots \xrightarrow{R_l} A_{l+1}$, where $A_i \in \mathcal{T}_\mathcal{A}$ and $R_i \in \mathcal{T}_\mathcal{E}$. This structure represents the sequence of meta-relations $(A_1, R_1, A_2)$, $(A_2, R_2, A_3)$, $\cdots$, $(A_l, R_l, A_{l+1})$.
Meta-paths serve as a powerful abstraction to capture complex, higher order semantics and structural patterns within a heterogeneous graph.

\subsection{Generative SSL on Heterogeneous Graphs}
Given a heterogeneous graph $\mathcal{G}$, an encoder $f_E$ and a decoder $f_D$. The task is to first learn the node embeddings $\mathbf{H}=f_E(\mathcal{G})$ of the graph using the encoder $f_E$, which encodes the rich structural and semantic information within the heterogeneous graph $\mathcal{G}$.
Then, the decoder $f_D$ is used to reconstruct some parts of the original graph, indicated as $\mathcal{G}^\prime=f_D(\mathbf{H})$, where $\mathcal{G}^\prime$ can be either reconstructed node attributes, adjacency matrix, or other attributes in the graph.
The learned embeddings $\mathbf{H}$ can be further utilized in various downstream graph mining tasks, such as node classification, node clustering, and link prediction.
Unlike contrastive learning, generative SSL is more straightforward and does not require complex data augmentation and negative sampling strategies.

\section{Methodology}

\begin{figure*}[t]
    \centering
    \includegraphics[width=2\columnwidth]{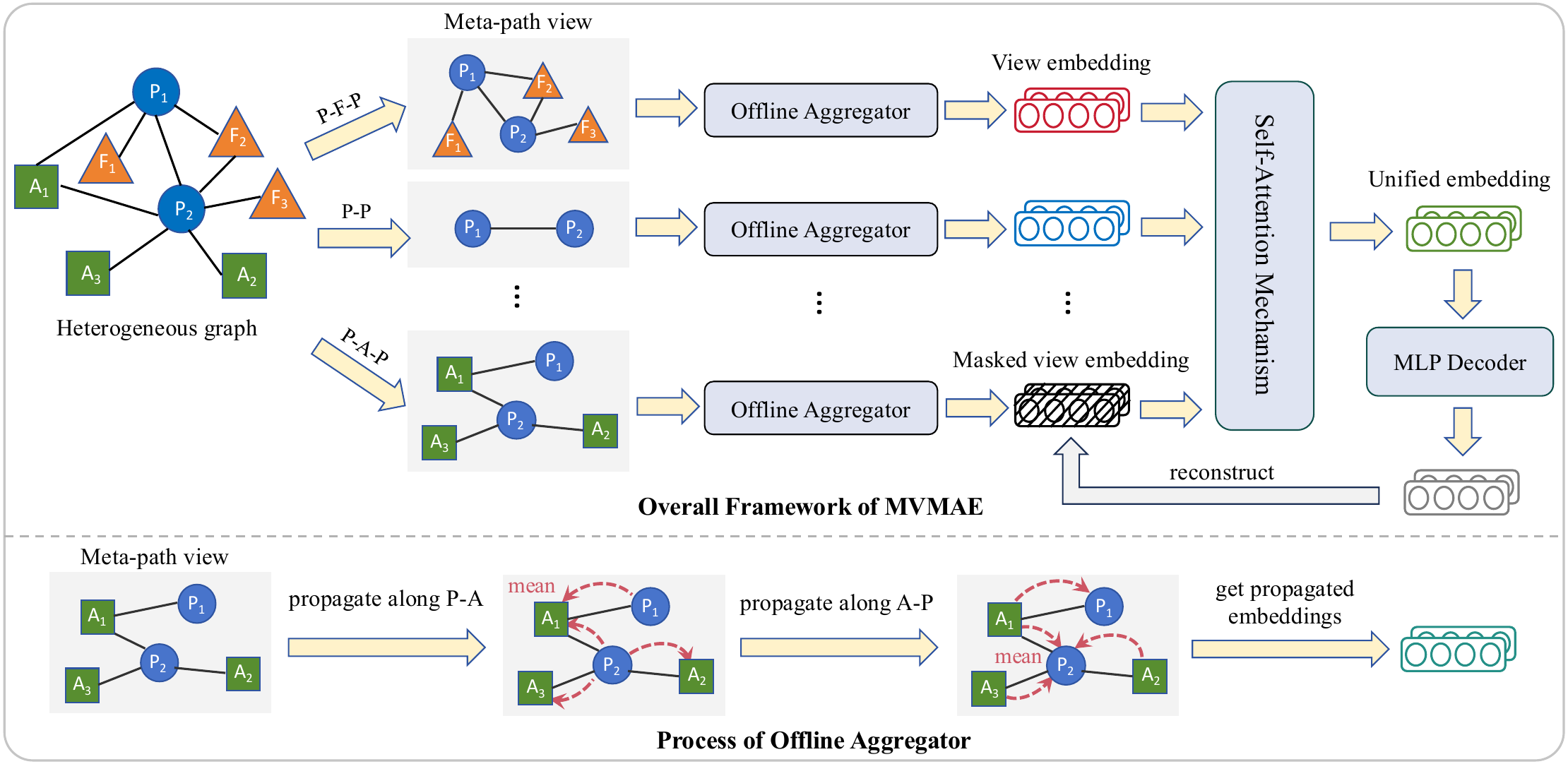}
        \caption{
    Framework of our proposed MVMAE model.
    The example is based on an academic heterogeneous graph with node types paper (P), author (A), and field (F). The figure demonstrates the aggregation of 1-hop meta-path P-P, and 2-hop meta-paths P-F-P and P-A-P.
    }
    \label{fig:model}
\end{figure*}

To address the challenges outlined in Section \ref{sec:intro}, we propose the multiview masked autoencoder (MVMAE)---a novel, efficient, and scalable framework for heterogeneous graph SSL.
The overall framework of our MVMAE model is illustrated in Fig. \ref{fig:model}, and the pseudocode is elaborated in Algorithm \ref{algorithm}.

\subsection{Overall Framework}

To address challenge C1, our MVMAE framework utilizes an efficient offline aggregation mechanism that precomputes meta-path-based neighbor representations in a preprocessing step, thereby conserving memory cost and eliminating redundant calculations during training.
To address challenges C2 and C3, we introduce a meta-path view-based mask-then-reconstruct strategy.
Unlike conventional methods that mask node features, we mask a portion of the meta-path view embeddings.
After masking, the view embeddings of each node are modeled as a sequence of tokens and fed into a Transformer encoder to generate a unified representation.
The fused embedding is subsequently used by a simple MLP decoder to reconstruct the content of the masked views.
By compelling the model to reconstruct masked information from visible views, this process ensures that the learned embeddings capture deep semantic correlations across the entire graph, resulting in highly expressive and robust representations.

We next describe each component of our model in detail.

\subsection{Efficient Offline Aggregation Mechanism}

To model meta-paths within heterogeneous graphs, many existing works \cite{DBLP:conf/aaai/0001DZZC23,DBLP:conf/aaai/ParkK0Y20,DBLP:conf/www/WangJSWYCY19} rely on extracting meta-path-induced subgraphs. However, this approach is computationally expensive and scales poorly to large graphs.
For instance, consider the P-A-P meta-path in an academic network. The corresponding induced subgraph, $\mathcal{G}_{\text{PAP}}$, is represented by an adjacency matrix $\mathbf{A}_{\text{PAP}}$, where each element $(\mathbf{A}_{\text{PAP}})_{ij}$ quantifies the number of paths connecting paper nodes $i$ and $j$ via the meta-path. The construction of $\mathbf{A}_{\text{PAP}}$ requires checking each pair of $(i, j)$, leading to a computational complexity of $O(n^2)$, which is prohibitive for large graphs.

Recent studies \cite{DBLP:conf/aaai/YangYPYF23,DBLP:conf/icml/WuSZFYW19} have demonstrated that sophisticated neighbor attention mechanisms are not essential for graph convolution, as simple mean aggregation can achieve comparable performance.
A notable example is the Simplified Graph Convolution (SGC) \cite{DBLP:conf/icml/WuSZFYW19}, which formulates the aggregation process as
\begin{equation} \label{equ:sgc}
\begin{gathered}
\tilde{\mathbf{A}} = \mathbf{D}^{-\frac{1}{2}} \mathbf{A} \mathbf{D}^{-\frac{1}{2}} \\
\mathbf{H}^{(k)} \leftarrow \tilde{\mathbf{A}} {\mathbf{H}}^{(k-1)} \Theta^{(k)}
\end{gathered}
\end{equation}
where $\mathbf{A} \in \mathbb{R}^{n \times n}$ is the adjacency matrix, $\mathbf{D}$ is the diagonal degree matrix, $\mathbf{H}^{(k)} \in \mathbb{R}^{n \times d}$ is the aggregated feature matrix at the $k$th layer, and $\Theta^{(k)} \in \mathbb{R}^{d \times d}$ is a layer-specific trainable weight matrix.

Inspired by the simplification in (\ref{equ:sgc}), we propose an efficient offline aggregation strategy for meta-path views, which avoids explicitly constructing the meta-path-induced subgraph such as $\mathcal{G}_{\text{PAP}}$.

To obtain the aggregated features of paper nodes via P-A-P meta-path, we perform the message propagation process as
\begin{equation}
\begin{gathered}
    \bar{\mathbf{X}}_{\text{A}} = \tilde{\mathbf{A}}_{\text{AP}} \mathbf{X}_{\text{P}} \\
    \bar{\mathbf{X}}_{\text{P}} = \tilde{\mathbf{A}}_{\text{PA}} \bar{\mathbf{X}}_{\text{A}}
\end{gathered}
\end{equation}
where $\mathbf{X}_{\text{P}}$ is the initial feature matrix of paper nodes, $\mathbf{A}_{\text{PA}}$ and $\mathbf{A}_{\text{AP}}$ are two adjacency matrices representing paper-author (P-A) and author-paper (A-P) relations, respectively, and $\tilde{\mathbf{A}}_{(\cdot)}$ denotes their normalized version.
The usage of the trainable weight matrix $\Theta$ is deferred to the training stage.
The final output $\bar{\mathbf{X}}_{\text{P}}$ represents these article features after aggregation across the meta-path.

The key points of our method lies in that $\mathbf{A}_{\text{PAP}}$ is actually constructed by $\mathbf{A}_{\text{PA}} \cdot \mathbf{A}_{\text{AP}}$, so we can rewrite the message propagation process as
\begin{equation} \label{eq:aggr1}
\bar{\mathbf{X}}_{\text{P}} = \tilde{\mathbf{A}}_{\text{PAP}} \mathbf{X}_{\text{P}} \Theta = \tilde{\mathbf{A}}_{\text{PA}} \tilde{\mathbf{A}}_{\text{AP}} \mathbf{X}_{\text{P}} \Theta.
\end{equation}
This computational strategy offers two main advantages.
First, the matrix multiplication can be performed sequentially, by first computing $\tilde{\mathbf{A}}_{\text{AP}} \mathbf{X}_{\text{P}}$, and then multiplying the result by $\tilde{\mathbf{A}}_{\text{PA}}$.
Given that real-world graphs are typically sparse, the matrices $\tilde{\mathbf{A}}_{\text{PA}}$ and $\tilde{\mathbf{A}}_{\text{AP}}$ can be stored in sparse formats.
Consequently, the overall complexity of (\ref{eq:aggr1}) is linear with respect to the number of edges.
Second, the computation of $\tilde{\mathbf{A}}_{\text{PA}} \tilde{\mathbf{A}}_{\text{AP}} \mathbf{X}_{\text{P}}$ involves no trainable parameters. This allows it to be precomputed during a preprocessing step and reused in each training epoch, substantially reducing computational cost.

Without loss of generality, we extend this procedure to arbitrary meta-paths.
Given a $l$-hop meta-path $\phi_m = c_0 c_1 c_2 \cdots c_l$ and node feature matrix $\mathbf{X}_{c_l}$ of node type $c_l$, the aggregation process across $\phi_m$ is defined as
\begin{equation}
    \label{eq:xphi}
    \mathbf{X}^{\phi_m} = \tilde{\mathbf{A}}_{c_0, c_1} \tilde{\mathbf{A}}_{c_1, c_2} \cdots \tilde{\mathbf{A}}_{c_{l-1}, c_l} \mathbf{X}_{c_l},
\end{equation}
where each $\hat{\mathcal{A}}_{c_i, c_j}$ represents the normalized adjacency matrix connecting nodes of type $c_i$ to those of type $c_j$.
The aggregated node embedding matrix $\mathbf{X}^{\phi_m}$ encapsulates the structural and semantic information within the meta-path $\phi_m$.

During each training epoch, we reuse the precomputed aggregation embeddings $\mathbf{X}^{\phi_m}$. Since the embeddings derived from different meta-paths may reside in distinct vector spaces, we project them into a unified latent space. This projection is achieved by applying a semantic-specific matrix $\mathbf{W}^{\phi_m}$ for each meta-path $\phi_m$, yielding the transformed features as
\begin{equation}
\mathbf{X}'^{\phi_m} = \mathbf{X}^{\phi_m} \mathbf{W}^{\phi_m}.
\end{equation}
The trainable projection matrix $\mathbf{W}^{\phi_m}$ is corresponding to the weight matrix $\Theta$ introduced in (\ref{equ:sgc}).

\subsection{Multiview Mask-Then-Reconstruct Strategy}

Another key innovation of our method is a refined masking strategy tailored for the MAE paradigm on heterogeneous graphs.
In conventional graph SSL methods \cite{DBLP:conf/kdd/HouLCDYW022,DBLP:conf/aaai/0001DZZC23}, the standard autoencoder framework involves masking a portion of the raw input features and train an encoder-decoder architecture to reconstruct them.
We contend that this approach is suboptimal for heterogeneous graphs for two primary reasons. First, raw node features are often sparse, incomplete, or entirely absent. Second, and more critically, these features inherently lack the rich structural and semantic information embedded within the heterogeneous graph's topology.
Consequently, reconstructing them serves as an insufficient learning objective, failing to adequately capture the complex internode relationships.

To overcome this limitation, we propose a novel mask-then-reconstruct strategy.
Instead of reconstructing raw features, our model is trained to predict the representations of masked meta-path views, leveraging them as a more informative and robust self-supervision signal.

In particular, during each training epoch, we first randomly sample a subset of meta-path views $\tilde{\Phi} \subset \Phi$ for masking, where $\Phi$ denotes the complete set of meta-path views and the size of the subset is controlled by the mask ratio $\rho$.
The initial representations corresponding to these selected views are then replaced by a learnable [MASK] token. This masking operation is formally defined as
\begin{equation} \label{eq:case}
\mathbf{H}^{\phi_m} =
\begin{cases}
\text{[MASK]}, & \text{if } \phi_m \in \tilde{\Phi} \\
\mathbf{X}'^{\phi_m}, & \text{if } \phi_m \notin \tilde{\Phi}
\end{cases},
\end{equation}
where the $\text{[MASK]}$ token is a learnable vector initialized by zeros.

Subsequently, the complete set of both masked and unmasked view representations is fed into a graph encoder (detailed in the next subsection).
The encoder fuses these varied inputs into a unified embedding for each node.
The unified embedding is then passed to a decoder, which is trained to reconstruct the original representations $\mathbf{X}^{\phi_m}$ for only the masked views, thus forming the self-supervision task.

A notable advantage of our meta-path view-based mask-then-reconstruct strategy is its effectiveness on graphs that lack initial node features.
For instance, on the AMiner dataset, node attributes are often initialized from a random Gaussian distribution, thus carrying no semantic information. Conventional MAE methods, which are designed to reconstruct these features, are rendered ineffective in this setting, since their learning objective degenerates into reconstructing random noise, providing no valuable supervisory signal.
On the contrary, our approach remains powerful as it learns from the graph's structural topology via meta-path views. In Section~\ref{sec:experiment}, our experimental results demonstrate that our proposed method achieves strong performance even in the absence of informative features.

\subsection{Multiview Fusion Encoder}

The multiview fusion encoder is designed to integrate semantic information from various meta-path views, producing a unified and high-quality node representation suitable for downstream tasks like node classification.

Drawing inspiration from the success of the Transformer architecture in sequence modeling \cite{DBLP:conf/nips/VaswaniSPUJGKP17}, we model each node as a sequence.
The sequence of a given node is composed of its representations under different meta-path views, with each view's representation acting as a token.
In particular, for each node $v$, we construct an input sequence by collecting its view-specific representations as $[\boldsymbol{h}_v^{\phi_1}, \boldsymbol{h}_v^{\phi_2}, \cdots, \boldsymbol{h}_v^{\phi_{|\Phi|}}]$, where each token $\boldsymbol{h}_v^{\phi_i}$ is extracted from the corresponding embedding matrix $\mathbf{H}_{\phi_i}$.

The multiview fusion encoder then employs a self-attention mechanism to model the intricate correlations and interdependencies among these view-based tokens.
By aggregating the token embeddings according to their learned attention weights, the model produces a single and comprehensive representation for the node, denoted as $\boldsymbol{z}$. This encoding process is formalized as
\begin{equation} \label{eq:transformer}
\begin{gathered}
    \boldsymbol{q}^{\phi_m} = \mathbf{W_Q} \boldsymbol{h}^{\phi_m}, \boldsymbol{k}^{\phi_m} = \mathbf{W_K} \boldsymbol{h}^{\phi_m} \\
    \boldsymbol{v}^{\phi_m} = \mathbf{W_V} \boldsymbol{h}^{\phi_m}, \phi_m \in \Phi \\
    \operatorname{Att}(\phi_m, \phi_n) = \boldsymbol{q}^{\phi_m} \cdot (\boldsymbol{k}^{\phi_n})^\top \\
    \alpha(\phi_m, \phi_n) = \frac{\exp(\operatorname{Att}(\phi_m, \phi_n))}{\sum_{\phi_k \in \Phi} \exp(\operatorname{Att}(\phi_m, \phi_k))} \\
    \boldsymbol{z}^{\phi_m} = \sum_{\phi_n \in \Phi} \alpha(\phi_m, \phi_n) \boldsymbol{v}^{\phi_n} + \boldsymbol{h}^{\phi_m},
\end{gathered}
\end{equation}
where $\mathbf{W_Q}$, $\mathbf{W_K}$, and $\mathbf{W_V}$ are trainable parameters shared for all meta-paths.
For brevity, we omit the subscript of node index $i$.
With the set of context-aware view embeddings $\{ \boldsymbol{z}^{\phi_m} \}_{\phi_m \in \Phi}$ obtained from the self-attention layer, we then aggregate them to produce the final and unified representation for the node.
This is achieved by concatenating all view-specific output vectors, followed by an MLP projection as
\begin{equation} \label{eq:z}
    \boldsymbol{z} = \operatorname{MLP}([\boldsymbol{z}^{\phi_1} || \boldsymbol{z}^{\phi_2} || \cdots || \boldsymbol{z}^{\phi_{\lvert \Phi \rvert}}]).
\end{equation}

\subsection{Simple Decoder}

A key aspect of our architecture is the deliberate simplicity of the decoder. In contrast to existing methods that employ powerful GNN models as decoders \cite{DBLP:conf/kdd/HouLCDYW022,DBLP:conf/aaai/0001DZZC23}, we utilize a simple MLP.
This design choice ensures that the encoder must generate highly informative representations, as it cannot rely on a complex decoder to compensate for information loss. This forces the learned embedding $\boldsymbol z$ to capture richer semantics of the heterogeneous graph.

The MLP decoder takes the final node embeddings $\boldsymbol z$ from (\ref{eq:z}) as input to predict the masked view representations $\mathbf{X}^{\phi_{\text{mask}}}$.
Since each meta-path possesses distinct semantics, we employ a separate decoder $f_D^{\phi_{\text{mask}}}$ for each target masked meta-path as
\begin{equation} \label{eq:decoder}
\boldsymbol{p}^{\phi_{\text{mask}}} = f_D^{\phi_{\text{mask}}}(\boldsymbol{z}) = \operatorname{MLP}^{\phi_{\text{mask}}}(\boldsymbol{z}),
\end{equation}
where $\boldsymbol{p}^{\phi_{\text{mask}}}$ is the reconstructed representations of the masked view.
This design is computationally efficient, as the overhead of the MLP decoders is minimal.

\subsection{Training and Evaluation}

For self-supervised training, the objective is to minimize the discrepancy between the reconstructed representations of a masked view $\boldsymbol{p}^{\phi_{\text{mask}}}$ and their corresponding ground-truth representations $\boldsymbol{x}^{\phi_{\text{mask}}}$.
Following prior works \cite{DBLP:conf/aaai/0001DZZC23,DBLP:conf/kdd/HouLCDYW022}, we formulate the reconstruction loss using a scaled cosine error as:
\begin{equation} \label{eq:loss}
    \mathcal{L}^{\phi_{\text{mask}}} = \frac{1}{\lvert \mathcal{V} \rvert} \sum_{v_i \in \mathcal{V}} (1 - \frac{(\boldsymbol{p}_i^{\phi_{\text{mask}}})^\top \cdot \boldsymbol{x}_i^{\phi_{\text{mask}}}}{\left\| \boldsymbol{p}_i^{\phi_{\text{mask}}} \right\| \cdot \left\| \boldsymbol{x}_i^{\phi_{\text{mask}}} \right\|})^\gamma, \gamma \geq 1,
\end{equation}
where $\gamma$ is a scaling exponent that sharpens the focus on harder-to-reconstruct samples.

To evaluate the quality of the learned representations, we adopt the linear probing protocol that is standard in graph SSL. This two-stage process works as follows:
First, the final node embeddings $\boldsymbol z$ are generated by our model after the self-supervised training phase. These embeddings are then frozen.
Second, a simple linear classifier (e.g., a single-layer MLP) is trained on top of these frozen embeddings to perform the downstream node classification task.
The final classification metrics are reported based on the performance of this linear classifier.

\subsection{Time and Space Complexity Analysis}

Our proposed MVMAE framework is designed for efficiency and scalability. To analyze its computational complexity, we separate the process into two key phases: a one-time offline preprocessing stage and the per-epoch online training.
Let $n$ denote the number of target nodes, $m$ denote the number of meta-paths, $d$ denote the embedding dimension, $l_{max}$ denote the maximum length of a meta-path, and $e$ denote the average number of edges of each relation.

\subsubsection{Time Complexity}
In the offline preprocessing phase, our efficient offline aggregation mechanism computes the aggregated features $\mathbf{X}^{\phi_m}$ for each meta-path $\phi_m$.
This is achieved through a sequence of sparse-dense matrix multiplications. For a meta-path of length $l$, this requires $l-1$ such multiplications, where the complexity of a single multiplication is linear in the number of edges.
Consequently, the total one-time preprocessing cost for all $m$ meta-paths is approximately $\mathcal{O}(m \cdot l_{max} \cdot e \cdot d)$. Since this computation is performed only once and offline, it does not affect the training speed.

In each online training epoch, it begins with view projection, where the $m$ precomputed embeddings are projected into a unified latent space, incurring a cost of $\mathcal{O}(m \cdot n \cdot d^2)$.
Next, the multiview fusion encoder, i.e., a Transformer with $L_{enc}$ layers, processes a sequence of $m$ tokens for each of the $n$ nodes. Its complexity is dominated by the self-attention mechanism, resulting in a cost of $\mathcal{O}(n \cdot L_{enc} \cdot (m^2 d + m d^2))$.
Finally, the decoding process, involving a simple MLP, reconstructs the representation for the masked view with a complexity of $\mathcal{O}(n \cdot d^2)$.
Combining these steps, the overall time complexity per epoch is $\mathcal{O}(mnd^2 + n L_{enc}(m^2 d + md^2))$.
Consequently, the complexity is effectively linear with respect to the number of nodes $n$, as other coefficients are comparatively small, which ensures the model's scalability to large-scale graphs.

\subsubsection{Space Complexity}
The framework's space complexity is mainly affected by the storage of model parameters and the precomputed embeddings.
The most significant memory requirement comes from storing the aggregated features $\mathbf{X}^{\phi_m}$ for all $m$ meta-path views, which occupies $\mathcal{O}(m \cdot n \cdot d)$ space.
The model parameters themselves, including the $m$ projection matrices, the multiview fusion encoder, and the $m$ decoders, require $\mathcal{O}((m+L_{enc})d^2)$ space.
Since $n$ is substantially larger than $d$ and $m$, the storage of precomputed embeddings becomes the dominant factor. This leads to an overall space complexity of $\mathcal{O}(m \cdot n \cdot d)$.

\begin{algorithm}[!tb]
\caption{Proposed MVMAE Framework}
\label{algorithm}
\flushleft{\textbf{Input}: Heterogeneous graph $\mathcal{G}$, initial node features $\mathbf{X}$, set of meta-paths $\Phi$}, mask ratio $\rho$ \\
\textbf{Output}: The trained encoder $f_E$

\begin{algorithmic}[1] \STATE \textbf{// Phase 1: One-time Offline Precomputation}
\FOR{each meta-path $\phi_m \in \Phi$}
    \STATE Compute aggregated features $\mathbf{X}^{\phi_m}$ via sequential sparse-dense matrix multiplications as in Equation~(\ref{eq:xphi}).
\ENDFOR

\STATE \textbf{// Phase 2: Online Training}
\STATE Initialize a learnable mask token embedding $[\text{MASK}]$.
\WHILE{not converged}
    \STATE // \textit{Project views into unified latent space}
    \FOR{each meta-path $\phi_m \in \Phi$}
        \STATE $\mathbf{X}'^{\phi_m} \leftarrow \mathbf{X}^{\phi_m} \mathbf{W}^{\phi_m}$
    \ENDFOR

    \STATE // \textit{Apply meta-path view masking}
    \STATE Randomly sample a set of views to mask: $\Phi_{\text{mask}} \subset \Phi$ where $|\Phi_{\text{mask}}| = \lfloor \rho \cdot |\Phi| \rfloor$.
    \STATE Let $\Phi_{\text{vis}} = \Phi \setminus \Phi_{\text{mask}}$ be the set of visible views.
    \STATE Construct input view representations $\{\mathbf{H}^{\phi_m}\}_{\phi_m \in \Phi}$ for the encoder as in Equation~(\ref{eq:case}):
    \STATE $\mathbf{H}^{\phi_m} \leftarrow \mathbf{X}'^{\phi_m}$ if $\phi_m \in \Phi_{\text{vis}}$
    \STATE $\mathbf{H}^{\phi_m} \leftarrow [\text{MASK}]$ if $\phi_m \in \Phi_{\text{mask}}$

    \STATE // \textit{Encoding and Decoding processes}
    \STATE Use multiview fusion encoder to generate unified node embeddings $\mathbf{Z} \leftarrow f_{E}(\{\mathbf{H}^{\phi_m}\}_{\phi_m \in \Phi})$ as in Equation~(\ref{eq:transformer}).
    \STATE Initialize total loss $\mathcal{L}_{total} \leftarrow 0$.
    \FOR{each masked view $\phi_{mask} \in \Phi_{\text{mask}}$}
        \STATE Reconstruct the masked view $\mathbf{P}^{\phi_{mask}} \leftarrow f_{D}^{\phi_{mask}}(\mathbf{Z})$ as in Equation~(\ref{eq:decoder})
        \STATE Compute reconstruction loss $\mathcal{L}^{\phi_{mask}}$ between $\mathbf{P}^{\phi_{mask}}$ and $\mathbf{X}^{\phi_{mask}}$ as in Equation~(\ref{eq:loss}).
        \STATE $\mathcal{L}_{total} \leftarrow \mathcal{L}_{total} + \mathcal{L}^{\phi_{mask}}$
    \ENDFOR

    \STATE Perform backpropagation on $\mathcal{L}_{total}$ and update all trainable parameters.
\ENDWHILE
\STATE \textbf{return} $f_{E}$
\end{algorithmic}
\end{algorithm}

\section{Experiments} \label{sec:experiment}

\begin{table*}[t!]
    \centering
    \caption{Statistics of the datasets.}
    \label{tab:statistics}
    \begin{tabular}{c|c|c|c|c|c}
    \hline
    Datasets & Node Types & \# Nodes & Edge Types & \# Edges & Meta-paths \\
    \hline
    \hline
    \multirow{3}{*}{ACM} & Paper & 4,019 & PA/AP & 13,407 & P-A-P, P-S-P, \\
    & Author & 7,167 & PS/SP & 4,019 & P-A-P-A-P, P-A-P-S-P, \\
    & Subject & 60 & & & P-S-P-A-P, P-S-P-S-P \\
    \hline
    \multirow{4}{*}{DBLP} & Author & 4,057 & AP/PA & 19,645 & A-P-A, \\
    & Paper & 14,328 & PT/TP & 85,810 & A-P-T-P-A, \\
    & Term & 7,723 & PC/CP & 14,328 & A-P-C-P-A \\
    & Conference & 20 & & & \\
    \hline
    \multirow{3}{*}{Freebase} & Movie & 4,278 & MD/DM & 4,278 & M-D-M, M-A-M, \\
    & Director & 2,081 & MA/AM & 12,828 & M-D-M-D-M, M-D-M-A-M, \\
    & Actor & 5,257 & & & M-A-M-D-M, M-A-M-A-M \\
    \hline
    \multirow{4}{*}{AMiner} & Paper & 127,623 & PA/AP & 355,072 & P-A-P, P-R-P, P-C-P, \\
    & Reference & 147,251 & PR/RP & 392,519 & P-A-P-A-P\, \\
    & Author & 164,473 & PC/CP & 127,623 & P-R-P-R-P, \\
    & Conference & 101 & & & P-C-P-C-P \\
    \hline
    \multirow{3}{*}{OAG} & Paper & 542,791 & PP & 5,722,208 & P-P-P, P-A-P, P-V-P, \\
    & Author & 216,963 & PA/AP & 1,171,274 & P-A-P-A-P, P-A-P-V-P, \\
    & Venue & 5,160 & PV/VP & 518,525 & P-V-P-A-P, P-V-P-V-P \\
    \hline
    \multirow{3}{*}{RCDD} & Item & 11,933,366 & IB/BI & 3,781,140 & I-B-I, I-F-I, \\
    & B & 45,602 & IF/FI & 11,683,940 & I-B-I-B-I, I-F-I-B-I, \\
    & F & 1,063,739 & & & I-B-I-F-I, I-F-I-F-I \\
    \hline
    \end{tabular}
\end{table*}

We evaluate the effectiveness and scalability of our proposed MVMAE model through extensive experiments on six real-world datasets, including three large-scale graphs.
We first perform a node classification task to measure overall performance, followed by a series of ablation studies to investigate the contribution of our model's key components.

\subsection{Experimental Setup}

\subsubsection{Datasets}

We evaluate our model on five publicly available heterogeneous graph benchmark datasets: ACM, DBLP, Freebase, AMiner, and OAG. These datasets have been widely adopted in previous works \cite{DBLP:conf/nips/0002GW0XLH22,DBLP:conf/aaai/0001DZZC23,DBLP:conf/kdd/WangLHS21,DBLP:conf/kdd/DuanXL24}. For a fair comparison, we adopt the same training, validation, and test set splits as the baseline model, HGMAE \cite{DBLP:conf/aaai/0001DZZC23}. Detailed statistics for each dataset are summarized in Table \ref{tab:statistics}.

\paragraph{ACM} The ACM\footnote{\url{https://www.acm.org}} dataset is extracted from the ACM digital library, which contains three node types: Paper (P), Author (A), and Subject (S), and four edge relations between them. Paper nodes are classified into three classes: Data Mining, Database, and Computer Network.

\paragraph{DBLP} The DBLP\footnote{\url{https://www.aminer.cn/citation}} dataset is extracted from the DBLP bibliography, which contains four node types: Author (A), Paper (P), Conference (C), and Term (T), and six edge relations between them. Author (A) nodes are classified into four research areas: Data Mining, Information Retrieval, Database, and Artificial Intelligence.

\paragraph{Freebase} The Freebase\footnote{\url{https://www.freebase.com}} dataset is extracted from the online movie rating website Freebase, which contains four node types: Movie (M), Actor (A), and Director (D), and six edge relations between them. Movie (M) nodes are classified into four classes by their genres: Comedy, Documentary, Drama, and Horror.

\paragraph{AMiner} The AMiner\footnote{\url{https://www.aminer.cn}} dataset is a large-scale bibliographic network containing three types of nodes: Paper (P), Author (A), and Reference (R), with four edge relations connecting them. In this dataset, Paper (P) nodes are classified into four distinct research fields: Computer Vision, Data Mining, Medical Informatics, and Theory.

\paragraph{OAG} The OAG\footnote{\url{https://open.aminer.cn/open/article?id=65bf053091c938e5025a31e2}} (Open Academic Graph) dataset is a large-scale academic graph that includes three primary node types: Paper (P), Author (A), and Venue (V), linked by five types of relations. The task is to classify Paper (P) nodes into ten different research areas.

{
\paragraph{RCDD} RCDD is a risk commodity detection dataset. Its node types consist of item (I), objects named ``f'' (F), objects named ``b'' (B), and others.
}

\subsubsection{Baselines}

We compare our proposed model MVMAE against nine representative baseline methods for heterogeneous graphs. These baselines are categorized into two groups.
The first group consists of three semisupervised methods: the classic HGNNs HAN \cite{DBLP:conf/www/WangJSWYCY19} and HGT \cite{DBLP:conf/www/HuDWS20}. The second group comprises six state-of-the-art self-supervised methods: DMGI \cite{DBLP:conf/aaai/ParkK0Y20}, HDGI \cite{DBLP:journals/corr/abs-1911-08538}, HeCo \cite{DBLP:conf/kdd/WangLHS21}, HGMAE \cite{DBLP:conf/aaai/0001DZZC23}, RMR \cite{DBLP:conf/kdd/DuanXL24}, MV-HetGNN \cite{DBLP:journals/tkde/ShaoXWWZZ23}, and Hetero2Net \cite{li2025heterophily}.

\subsubsection{Evaluation Metrics}

We evaluate the performance of our proposed MVMAE model and all baselines on a downstream node classification task, using Macro- and Micro-F1 scores as the primary evaluation metrics.

For supervised baselines, the models are trained end-to-end using a cross-entropy loss function, and their performance is subsequently reported on the test set.
For our self-supervised method MVMAE and the corresponding baselines, we follow a standard linear evaluation protocol. First, each model is pretrained using its respective self-supervised objective to generate node embeddings. Subsequently, a logistic regression classifier is trained on these frozen embeddings, and its performance on the test set is used to assess the quality of the learned representations.

\subsubsection{Implementation Details}

Our proposed MVMAE model employs a single-layer Transformer in the multiview fusion encoder with a single attention head to fuse the representation of meta-path views. The specific adopted meta-paths for each dataset are detailed in Table \ref{tab:statistics}. For hyperparameters, the masking ratio is selected in the range of 0.1 to 0.9, the hidden layer dimensionality is selected from the set \{64, 128, 256, 512\}, the dropout ratio is selected in the range of 0.0 to 0.5, and the learning rate and weight decay are selected in the range of 1e-4 to 1e-2. We perform a grid search on the validation set and report the best results.
All model parameters are initialized using the Xavier uniform distribution \cite{Glorot2010UnderstandingTD} and optimized with the Adam optimizer \cite{Kingma2014AdamAM}.

The model is pretrained for a maximum of 500 epochs, and we select the checkpoint that achieves the lowest validation loss.
After pretraining, the learned node embeddings are frozen and fed into a linear classifier for the downstream node classification task. The final reported metrics are derived from this evaluation.

For baseline methods, we reproduce their results, adhering to the hyperparameter configurations specified in their original papers and official source codes. To ensure robustness, all evaluations are repeated five times with different random seeds, and we report the average results. All experiments are conducted on an NVIDIA RTX 3090 Ti GPU.

{
\subsubsection{Meta-path Selection Heuristics}

Following the common practice in heterogeneous graph learning (e.g., HAN and HGMAE), we select meta-paths based on two criteria: 1) Domain Knowledge: We prioritize meta-paths that capture essential semantic relations for the specific task (e.g., coauthorship in DBLP). 2) Path Length: We typically choose short meta-paths (i.e., $length \le 4$) to mitigate the over-smoothing issue and ensure computational efficiency.
}

\subsection{Performance Evaluation}

In this section, we present the experimental results on both supervised and unsupervised downstream tasks to quantitatively demonstrate the effectiveness of our proposed model MVMAE.

\begin{table*}[ht]
\caption{Node classification results (\%).}
\label{tab:classify}
\centering
\tabcolsep=0.06cm
\resizebox{\textwidth}{!}{
\begin{tabular}{ccccccccccccc}
\toprule
Dataset & Metric & Split & HAN & HGT & DMGI & HDGI & HeCo & HGMAE & RMR & MV-HetGNN & Hetero2Net & MVMAE \\
\midrule
\multirow{6}[0]{*}{ACM} & \multirow{3}[0]{*}{Mic-F1}
&  20 & 85.11$\pm$2.2 & 86.23$\pm$2.1 & 87.60$\pm$0.8 & 86.53$\pm$1.4 & 88.13$\pm$0.8 & 90.24$\pm$0.5 & 87.58$\pm$0.5 & 89.33$\pm$0.4 & 88.70$\pm$0.2 & \best{92.46$\pm$0.4} \\
&& 40 & 87.21$\pm$1.2 & 87.46$\pm$1.3 & 86.02$\pm$0.9 & 85.88$\pm$1.2 & 87.45$\pm$0.5 & 90.18$\pm$0.6 & 87.73$\pm$0.7 & 91.33$\pm$0.8 & 90.60$\pm$0.8 & \best{91.53$\pm$0.9} \\
&& 60 & 88.10$\pm$1.2 & 88.05$\pm$1.1 & 87.82$\pm$0.5 & 87.10$\pm$1.3 & 88.71$\pm$0.5 & 91.34$\pm$0.4 & 88.10$\pm$0.6 & 89.63$\pm$0.9 & 89.90$\pm$0.2 & \best{92.20$\pm$0.2} \\
\cmidrule{2-13}
& \multirow{3}[0]{*}{Mac-F1}
&  20 & 85.66$\pm$2.1 & 86.47$\pm$1.9 & 87.86$\pm$0.2 & 86.83$\pm$0.9 & 88.56$\pm$0.8 & 90.66$\pm$0.4 & 87.63$\pm$0.5 & 89.18$\pm$0.6 & 88.56$\pm$0.2 & \best{92.36$\pm$0.5} \\
&& 40 & 87.47$\pm$1.1 & 87.48$\pm$1.4 & 86.23$\pm$0.8 & 86.27$\pm$0.8 & 87.61$\pm$0.5 & 90.15$\pm$0.6 & 88.92$\pm$0.4 & 91.34$\pm$0.8 & 90.57$\pm$0.8 & \best{91.51$\pm$0.9} \\
&& 60 & 88.41$\pm$1.1 & 88.53$\pm$1.3 & 87.97$\pm$0.4 & 87.14$\pm$0.5 & 89.04$\pm$0.5 & 91.59$\pm$0.4 & 90.33$\pm$0.5 & 89.69$\pm$0.8 & 89.97$\pm$0.2 & \best{92.31$\pm$0.2} \\
\midrule
\multirow{6}[0]{*}{DBLP} & \multirow{3}[0]{*}{Mic-F1}
&  20 & 90.16$\pm$0.9 & 90.89$\pm$1.1 & 90.78$\pm$0.3 & 89.46$\pm$0.8 & 91.97$\pm$0.2 & 92.71$\pm$0.5 & 92.95$\pm$0.4 & 86.56$\pm$0.6 & 90.43$\pm$0.6 & \best{94.96$\pm$0.2} \\
&& 40 & 89.47$\pm$0.9 & 90.03$\pm$1.0 & 89.92$\pm$0.4 & 89.53$\pm$0.5 & 90.76$\pm$0.3 & 92.43$\pm$0.3 & 93.12$\pm$0.4 & 87.80$\pm$1.1 & 91.63$\pm$0.1 & \best{93.70$\pm$0.2} \\
&& 60 & 90.34$\pm$0.8 & 91.11$\pm$0.9 & 90.66$\pm$0.5 & 89.91$\pm$0.6 & 91.59$\pm$0.2 & 93.05$\pm$0.3 & 93.11$\pm$0.3 & 91.20$\pm$1.0 & 93.50$\pm$0.1 & \best{95.76$\pm$0.0} \\
\cmidrule{2-13}
& \multirow{3}[0]{*}{Mac-F1}
&  20 & 89.31$\pm$0.9 & 90.77$\pm$0.5 & 89.94$\pm$0.4 & 88.37$\pm$0.7 & 91.28$\pm$0.2 & 92.28$\pm$0.5 & 92.86$\pm$0.4 & 85.51$\pm$0.6 & 89.80$\pm$0.6 & \best{94.47$\pm$0.2} \\
&& 40 & 88.87$\pm$1.0 & 89.24$\pm$0.8 & 89.25$\pm$0.4 & 88.53$\pm$0.8 & 90.34$\pm$0.3 & 92.12$\pm$0.3 & 93.13$\pm$0.3 & 87.22$\pm$1.1 & 91.20$\pm$0.1 & \best{93.40$\pm$0.2} \\
&& 60 & 89.20$\pm$0.8 & 89.25$\pm$0.9 & 89.46$\pm$0.6 & 89.01$\pm$0.6 & 90.64$\pm$0.3 & 92.33$\pm$0.3 & 93.10$\pm$0.3 & 90.30$\pm$1.2 & 92.73$\pm$0.1 & \best{95.20$\pm$0.0} \\
\midrule
\multirow{6}[0]{*}{Freebase} & \multirow{3}[0]{*}{Mic-F1}
&  20 & 65.66$\pm$0.8 & 63.20$\pm$0.4 & 52.46$\pm$1.5 & 62.43$\pm$0.8 & 65.96$\pm$0.6 & 66.03$\pm$0.6 & 58.06$\pm$0.8 & 55.80$\pm$0.0 & 57.36$\pm$1.6 & \best{66.40$\pm$0.5} \\
&& 40 & 68.10$\pm$0.4 & 66.10$\pm$0.4 & 53.40$\pm$1.4 & 64.60$\pm$0.6 & 65.40$\pm$0.2 & 65.30$\pm$0.2 & 58.26$\pm$1.0 & 56.70$\pm$0.0 & 58.23$\pm$1.1 & \best{68.80$\pm$1.0} \\
&& 60 & 67.80$\pm$0.7 & 64.53$\pm$0.1 & 52.70$\pm$0.7 & 62.30$\pm$1.3 & 65.46$\pm$0.9 & 65.15$\pm$0.3 & 61.46$\pm$0.6 & 55.46$\pm$0.1 & 61.53$\pm$0.8 & \best{68.53$\pm$0.4} \\
\cmidrule{2-13}
& \multirow{3}[0]{*}{Mac-F1}
& 20 & 54.82$\pm$5.1 & 54.82$\pm$2.7 & 49.01$\pm$0.8 & 46.33$\pm$0.9 & 54.04$\pm$3.4 & 55.88$\pm$5.5 & 45.93$\pm$1.4 & 48.30$\pm$0.0 & 51.29$\pm$5.3 & \best{58.72$\pm$1.7} \\
&& 40 & 60.68$\pm$2.1 & 59.22$\pm$0.1 & 52.40$\pm$4.3 & 51.19$\pm$3.8 & 48.16$\pm$0.1 & 47.84$\pm$0.1 & 47.27$\pm$3.6 & 57.14$\pm$0.0 & 59.58$\pm$4.9 & \best{61.13$\pm$2.3} \\
&& 60 & 58.12$\pm$2.0 & 53.46$\pm$2.2 & 52.84$\pm$5.0 & 45.78$\pm$1.4 & 47.91$\pm$0.7 & 56.34$\pm$2.3 & 48.41$\pm$0.3 & 57.74$\pm$3.4 & 59.23$\pm$2.1 & \best{61.47$\pm$1.0} \\
\midrule
\multirow{6}[0]{*}{AMiner} & \multirow{3}[0]{*}{Mic-F1}
&  20 & OOM & 89.70$\pm$0.1 & OOM & OOM & OOM & OOM & 93.01$\pm$0.6 & 89.52$\pm$9.5 & 90.42$\pm$0.5 & \best{96.10$\pm$0.5} \\
&& 40 & OOM & 91.43$\pm$0.2 & OOM & OOM & OOM & OOM & 94.43$\pm$0.5 & 91.67$\pm$0.8 & 92.27$\pm$0.7 & \best{98.35$\pm$0.4} \\
&& 60 & OOM & 92.25$\pm$0.1 & OOM & OOM & OOM & OOM & 95.27$\pm$0.4 & 95.62$\pm$0.5 & 94.36$\pm$0.5 & \best{98.52$\pm$0.4} \\
\cmidrule{2-13}
& \multirow{3}[0]{*}{Mac-F1}
&  20 & OOM & 92.27$\pm$0.1 & OOM & OOM & OOM & OOM & 90.81$\pm$0.9 & 88.67$\pm$8.2 & 86.62$\pm$0.9 & \best{94.03$\pm$0.6} \\
&& 40 & OOM & 94.37$\pm$0.4 & OOM & OOM & OOM & OOM & 93.11$\pm$0.5 & 92.09$\pm$0.3 & 92.45$\pm$1.1 & \best{95.96$\pm$0.5} \\
&& 60 & OOM & 94.89$\pm$0.3 & OOM & OOM & OOM & OOM & 94.30$\pm$0.4 & 94.70$\pm$0.3 & 93.83$\pm$0.8 & \best{96.48$\pm$0.4} \\
\midrule
\multirow{6}[0]{*}{OAG} & \multirow{3}[0]{*}{Mic-F1}
&  20 & OOM & 68.73$\pm$0.4 & OOM & OOM & OOM & OOM & 69.32$\pm$0.3 & 69.36$\pm$0.3 & 69.58$\pm$0.2 & \best{70.68$\pm$0.2} \\
&& 40 & OOM & 69.46$\pm$0.5 & OOM & OOM & OOM & OOM & 70.64$\pm$0.2 & 69.63$\pm$0.4 & 71.13$\pm$0.4 & \best{72.12$\pm$0.3} \\
&& 60 & OOM & 70.82$\pm$0.5 & OOM & OOM & OOM & OOM & 70.43$\pm$0.2 & 70.02$\pm$0.2 & 71.22$\pm$0.4 & \best{72.83$\pm$0.2} \\
\cmidrule{2-13}
& \multirow{3}[0]{*}{Mac-F1}
&  20 & OOM & 69.04$\pm$0.5 & OOM & OOM & OOM & OOM & 69.13$\pm$0.3 & 68.42$\pm$0.6 & 69.22$\pm$0.2 & \best{70.54$\pm$0.3} \\
&& 40 & OOM & 69.58$\pm$0.4 & OOM & OOM & OOM & OOM & 70.65$\pm$0.3 & 69.54$\pm$0.4 & 70.79$\pm$0.3 & \best{72.80$\pm$0.3} \\
&& 60 & OOM & 70.87$\pm$0.4 & OOM & OOM & OOM & OOM & 70.85$\pm$0.2 & 69.65$\pm$0.2 & 70.93$\pm$0.2 & \best{72.51$\pm$0.2} \\
\bottomrule
\end{tabular}
}
\end{table*}

\begin{table}[ht]
\caption{Node classification results on the RCDD dataset (\%).}
\label{tab:rcdd}
\centering
\resizebox{0.46\textwidth}{!}{
\begin{tabular}{cccc}
\toprule
Method & Micro-F1 & Macro-F1 & AP \\
\midrule
HAN & OOM & OOM & OOM \\
HGT & 98.25 ± 0.1 & 91.57 ± 0.2 & 92.29 ± 0.1 \\
DMGI & OOM & OOM & OOM \\
HDGI & OOM & OOM & OOM \\
HeCo & OOM & OOM & OOM \\
HGMAE & OOM & OOM & OOM \\
RMR & 98.03 ± 0.0 & 90.49 ± 0.1 & 88.82 ± 0.4 \\
MV-HetGNN & 98.31 ± 0.0 & 91.62 ± 0.1 & 89.67 ± 0.0 \\
Hetero2Net & \best{98.42 ± 0.0} & \best{92.60 ± 0.2} & \best{92.80 ± 0.0} \\
\midrule
MVMAE & 98.38 ± 0.0 & 92.14 ± 0.1 & 92.54 ± 0.0 \\
\bottomrule
\end{tabular}
}
\end{table}

\subsubsection{Node Classification}

To evaluate the quality of the learned node embeddings, we perform a downstream node classification task. Following the experimental setup of HGMAE \cite{DBLP:conf/aaai/0001DZZC23}, we use training sets of 20, 40, and 60 labeled nodes per class, with validation and test sets each containing 1,000 nodes.
For self-supervised models, which are pretrained without ground-truth labels, a linear logistic regression classifier is subsequently trained on the frozen node embeddings to obtain classification results.
For semisupervised methods, we directly report the classification results from their models. Performance is evaluated using Micro- and Macro-F1 scores.

As presented in Tables \ref{tab:classify} and \ref{tab:rcdd}, our proposed MVMAE model achieves state-of-the-art performance, outperforming all self-supervised baselines and even semisupervised ones.
We attribute the performance advantage of MVMAE over the strong baseline, HGMAE, to a fundamental difference in their pretraining strategies. Whereas HGMAE randomly masks individual nodes or edges, which can introduce sampling biases, MVMAE masks entire meta-path-based views. We suggest that this view-level masking strategy constitutes a more challenging and robust pretext task. This strategy makes the model less susceptible to local structural imbalances, enabling it to learn more global informative representations.

Furthermore, our experimental results highlight a significant trend that leading self-supervised methods are becoming highly competitive with, and in some cases superior to, their semisupervised counterparts.
This is directly evidenced by our MVMAE model, which outperforms all semisupervised baselines. Such a finding highlights the maturation of SSL and affirms its substantial potential for representation learning on heterogeneous graphs.

Notably, on the AMiner dataset, MVMAE demonstrates a particularly significant performance margin over other baselines. We attribute this to our novel multiview mask-and-reconstruct strategy. The AMiner dataset lacks initial node features, which we initialized from a random Gaussian distribution. Such random input features contain no semantic information.
In such a scenario, our strategy shifts the learning objective away from reconstructing raw features toward the higher-level task of inferring information across different structural views. This provides a robust supervisory signal that is effective even when initial node features are absent or flawed. Consequently, as the results show, our model excels and surpasses all other baselines on this dataset.

\subsubsection{Node Clustering}

{
Following the experimental setup in HGMAE \cite{DBLP:conf/aaai/0001DZZC23}, we compare our model with unsupervised baselines, as the node representation generation process for clustering is inherently unsupervised. The experimental results, as presented in Table~\ref{tab:cluster}, demonstrate that our proposed MVMAE consistently achieves the best performance across all four datasets, further validating the robustness and generalization ability of our model in unsupervised scenarios.
}

\begin{table}[ht]
\caption{Node clustering results (\%).}
\label{tab:cluster}
\centering
\resizebox{\columnwidth}{!}{
\begin{tabular}{ccccccccc}
\toprule
Dataset & \multicolumn{2}{c}{ACM} & \multicolumn{2}{c}{DBLP} & \multicolumn{2}{c}{Freebase} & \multicolumn{2}{c}{AMiner} \\
\cmidrule{1-9}
Metric & NMI & ARI & NMI & ARI & NMI & ARI & NMI & ARI \\
\midrule
DMGI & 46.53 & 42.16 & 47.35 & 46.28 & 19.72 & 19.56 & OOM & OOM \\
HDGI & 49.76 & 44.35 & 59.24 & 60.43 & 18.13 & 19.44 & OOM & OOM \\
HeCo & 56.17 & 56.36 & 74.71 & 79.41 & 21.33 & 22.99 & OOM & OOM \\
HGMAE & 71.20 & 74.33 & 75.23 & 80.63 & 19.91 & 21.04 & OOM & OOM \\
RMR & 51.86 & 47.27 & 9.90 & 7.21 & 13.08 & 11.82 & 27.42 & 18.82 \\
MVMAE & \best{72.81} & \best{76.32} & \best{77.50} & \best{83.17} & \best{21.60} & \best{23.18} & \best{36.67} & \best{24.12} \\
\bottomrule
\end{tabular}
}
\end{table}

\subsection{Sensitivity Analysis}

In this section, we present a sensitivity analysis to validate the robustness of our proposed MVMAE model. We conduct comprehensive ablation studies on its key components, investigate the impact of the masking ratio and various meta-path configurations, and compare its computational efficiency with baseline methods in terms of time and memory consumption.

\begin{table}
  \newcommand{\tabincell}[2]{\begin{tabular}{@{}#1@{}}#2\end{tabular}}
  \centering
  \caption{Ablation results of MVMAE on five datasets.}
    \label{tab:ablation}
  \resizebox{0.95\columnwidth}{!}{
    \begin{tabular}{c|c|c c c c c|c}
    \toprule
    Datasets & Metrics & w/o mask & w/o aggr & w/o fusion & MVMAE \\
    \midrule
    \multirow{2}{*}{ACM} & Mic-F1 & 90.24 & 91.07 & 88.73 & \best{92.46} \\
    & Mac-F1 & 90.16 & 91.33 & 89.76 & \best{92.36} \\
    \midrule
    \multirow{2}{*}{DBLP} & Mic-F1 & 93.12 & 94.37 & 92.68 & \best{94.96} \\
    & Mac-F1 & 93.02 & 94.16 & 92.23 & \best{94.47} \\
    \midrule
    \multirow{2}{*}{Freebase} & Mic-F1 & 62.76 & 63.92 & 60.44 & \best{66.40} \\
    & Mac-F1 & 55.66 & 56.15 & 54.27 & \best{58.72} \\
    \midrule
    \multirow{2}{*}{AMiner} & Mic-F1 & 95.53 & OOM & 93.31 & \best{96.10} \\
    & Mac-F1 & 93.69 & OOM & 92.18 & \best{94.03} \\
    \midrule
    \multirow{2}{*}{OAG} & Mic-F1 & 70.59 & OOM & 68.26 & \best{70.68} \\
    & Mac-F1 & 70.50 & OOM & 68.03 & \best{70.54} \\
    \bottomrule
    \end{tabular}
  }
\end{table}

\begin{table}
  \newcommand{\tabincell}[2]{\begin{tabular}{@{}#1@{}}#2\end{tabular}}
  \centering
  \caption{Further Ablation results of MVMAE on five datasets.}
    \label{tab:more_ablation}
  \resizebox{0.95\columnwidth}{!}{
    \begin{tabular}{c|c|c c c c c|c}
    \toprule
    Datasets & Metrics & Learnable Aggregator & Expressive Decoder & MVMAE \\
    \midrule
    \multirow{2}{*}{ACM} & Mic-F1 & \best{92.54} & 92.42 & 92.46 \\
     & Mac-F1 & 92.21 & \best{92.38} & 92.36 \\
    \midrule
    \multirow{2}{*}{DBLP} & Mic-F1 & 94.92 & 94.75 & \best{94.96} \\
     & Mac-F1 & 94.35 & 94.41 & \best{94.47} \\
    \midrule
    \multirow{2}{*}{Freebase} & Mic-F1 & 66.25 & 66.31 & \best{66.40} \\
     & Mac-F1 & 58.68 & \best{58.84} & 58.72 \\
    \midrule
    \multirow{2}{*}{AMiner} & Mic-F1 & 96.02 & 95.94 & \best{96.10} \\
     & Mac-F1 & \best{94.11} & 93.96 & 94.03 \\
    \midrule
    \multirow{2}{*}{OAG} & Mic-F1 & 70.61 & \best{70.75} & 70.68 \\
     & Mac-F1 & 70.49 & 70.51 & \best{70.54} \\
    \bottomrule
    \end{tabular}
  }
\end{table}

\begin{table*}[!t]
\centering
\caption{Efficiency comparison of different methods regarding parameter count (\# Params), memory footprint, and training time. Memory consumption is reported as the peak GPU memory usage in megabytes (MBs). Training time denotes the average duration per epoch in seconds (s), calculated over ten training epochs. All measurements are conducted on an NVIDIA GeForce RTX 3090 GPU (24-GB memory) and 128-GB RAM.}
\label{tab:time_memory}

\resizebox{0.85\textwidth}{!}{
\begin{tabular}{l|rrr|rrr|rrr}
\toprule
\multirow{2}{*}{Method} & \multicolumn{3}{c|}{ACM (N: 11,246)} & \multicolumn{3}{c|}{AMiner (N: 439,448)} & \multicolumn{3}{c}{OAG (N: 764,914)} \\ \cline{2-10}
& \# Params & Memory & Time & \# Params & Memory & Time & \# Params & Memory & Time \\
\hline
HAN & 394,499 & 15,526M & 0.30s & OOM & OOM & OOM & OOM & OOM & OOM \\
HGT & 462,231 & 806M & 0.27s & 728,114 & 4,568M & 2.3s & 1,679,676 & 23,940M & 5.7s \\
HGMAE & 285,404 & 7,624M & 0.32s & OOM & OOM & OOM & OOM & OOM & OOM \\
RMR & 85,216 & 762M & 0.23s & 92,352 & 3,052M & \best{0.2s} & 174,208 & 15,996M & \best{0.5s} \\
MV-HetGNN & 103,636 & 976M & 0.28s & 114,538 & 4,372M & 0.8s & 242,446 & 22,784M & 3.8s \\
Hetero2Net & 97,382 & 876M & 0.25s & 103,548 & 4,268M & 0.6s & 210,426 & 18,764M & 2.6s \\
\hline
MVMAE & \best{64,092} & \best{588M} & \best{0.21s} & \best{70,784} & \best{1,132M} & 0.8s & \best{166,208} & \best{7,826M} & 3.2s \\
\bottomrule
\end{tabular}}
\end{table*}

\subsubsection{Impact of Multiview Mask-Then-Reconstruct Strategy}

We perform an ablation study to validate the contribution of our proposed multiview mask-then-reconstruct strategy. In particular, we remove this component and adopt a random node feature masking strategy, following the precedent of GraphMAE \cite{DBLP:conf/kdd/HouLCDYW022} and HGMAE \cite{DBLP:conf/aaai/0001DZZC23}. The masking ratio is varied from 10\% to 80\% in increments of 10\%, and the best performance is reported.
In addition, the reconstruction objective is changed to the original node features instead of the view-based representations. The experimental results are reported in the ``w/o mask'' column of Tables \ref{tab:ablation} and \ref{tab:more_ablation}.

As shown in the ablation results, the random masking approach yields consistently lower performance than our proposed meta-path view masking on all five datasets. This limitation stems from the restricted semantic information inherent in the original node features.
This effect is most evident on the AMiner dataset, where the performance drop is the most significant. We attribute this to the fact that the AMiner dataset contains no original node features, so they are initialized from a random Gaussian distribution. Such random features have no semantic meaning.
Our method circumvents this issue by shifting the learning objective from the reconstruction of noise-like features to a more sophisticated task, e.g., inferring cross-view information from the graph's structure.
This innovation point generates a powerful supervisory signal, proving effective even in the absence of informative initial features.
As a result, our strategy substantially boosts performance in the challenging case of the AMiner dataset.

\subsubsection{Impact of Efficient Offline Aggregation Mechanism}

To validate the efficacy of our proposed offline message propagation mechanism, we conduct a comparative analysis against a learnable counterpart. In particular, we substitute our parameter-free approach with the widely adopted HAN \cite{DBLP:conf/www/WangJSWYCY19} to generate representations for each meta-path view. A critical implementation detail is that the reconstruction target was detached from the computational graph to prevent the model from learning trivial solutions.

The empirical results are summarized in the ``w/o aggr'' column of Table~\ref{tab:ablation}, which reports the performance of the HAN-based variant.
Our analysis reveals that our parameter-free mechanism consistently outperforms this learnable alternative on the majority of datasets.
Notably, the HAN-based model encounters out-of-memory (OOM) errors on large-scale graphs such as AMiner and OAG. On the contrary, our method not only executes successfully on these large graphs but also achieves highly competitive performance, underscoring its superior scalability and efficiency. Table~\ref{tab:time_memory} further supports this efficiency advantage, as MVMAE requires the fewest parameters and the lowest peak GPU memory across all three evaluated datasets, while maintaining competitive per-epoch training time.

\subsubsection{Impact of Multiview Fusion Encoder}

This study is conducted to validate the efficacy of our proposed multiview aggregation encoder. In this analysis, we remove the encoder and replace it with a naive averaging function to fuse the representations from each view. As detailed in the ``w/o fuse'' column of Table~\ref{tab:ablation}, this modification leads to a consistent degradation in node classification performance across all five datasets.

This anticipated outcome underscores the significant semantic heterogeneity among the various meta-path views. For instance, the meta-path P-A-P captures a strong, direct collaborative tie through coauthorship. On the contrary, the P-C-P view signifies a broader topical connection by linking papers presented at the same venue. Treating these semantically distinct views as equally important by simply averaging their representations can introduce a substantial bias, conflating their unique relational signals and ultimately generating a less informative final embedding.

\begin{figure}[t]
  \centering
  \includegraphics[width=1\columnwidth]{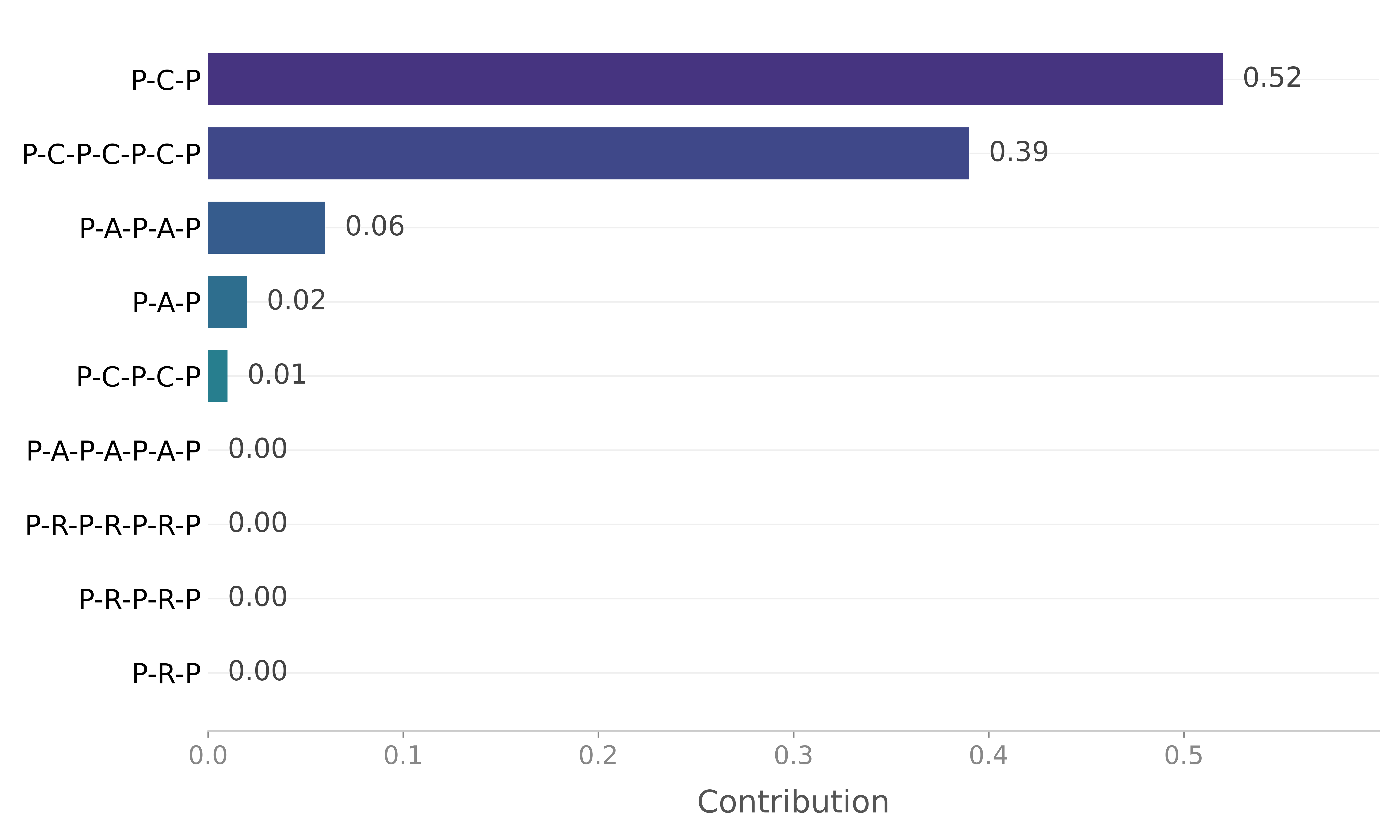}
    \caption{Sensitivity analysis of different meta-paths' contributions to the model performance on the AMiner dataset.}
  \label{fig:metapath_contribution}
    \end{figure}

\subsubsection{Contribution of Different Meta-Paths}

In this study, we quantitatively assess the contribution of individual meta-paths by examining the attention weights generated by our multiview fusion encoder. The learned weights, as visualized in Fig.~\ref{fig:metapath_contribution}, reveal a nonuniform contribution among the different meta-paths, signifying that the model selectively prioritizes more informative views.

Notably, on the AMiner dataset, the P-C-P meta-path consistently receives the highest attention score. This finding aligns with domain-specific expectations.
As academic conferences are typically organized around specific research topics, and the downstream task is academic field classification, the conference where a paper is published becomes a strong indicator of its field.
Therefore, the P-C-P structure, which connects papers presented at the same conference, provides a powerful and direct signal for identifying a paper's research field, making it a highly discriminative feature for the model to leverage.

\subsubsection{Impact of the masking ratio}

\begin{figure}[t]
  \centering
  \includegraphics[width=1\columnwidth]{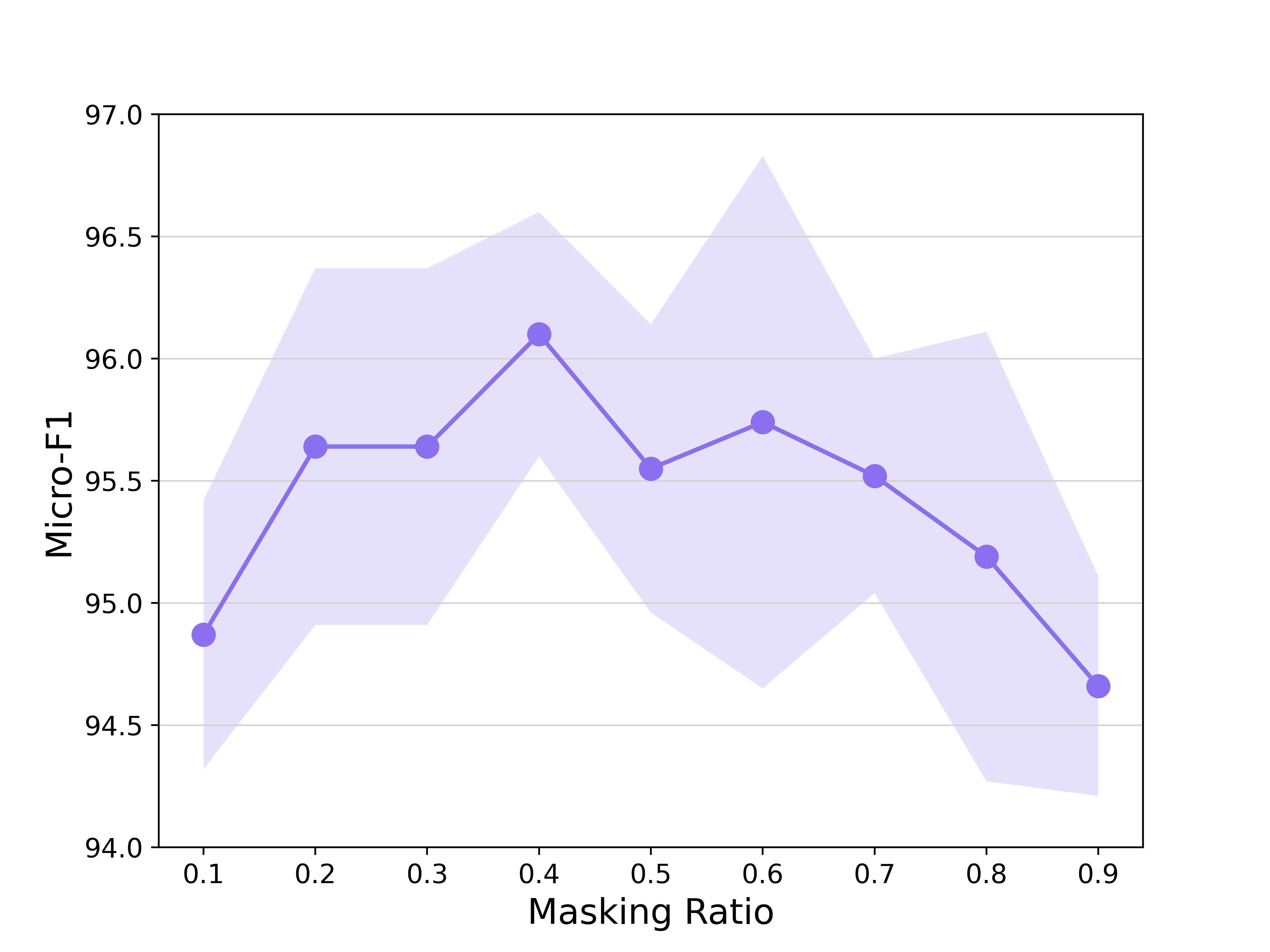}
    \caption{Sensitivity analysis of the masking ratio on the AMiner dataset.}
  \label{fig:mask_ratio}
    \end{figure}

In our approach, a portion of the meta-path view set for each node is randomly masked. To analyze the impact of this crucial hyperparameter, we conduct a sensitivity analysis by varying the masking ratio. The results, as presented in Fig. \ref{fig:mask_ratio}, demonstrate that the choice of masking ratio is critical to the model's performance, following a consistent trend across datasets. For example, on the AMiner dataset, a masking ratio of 0.4 achieves optimal performance.

This phenomenon illustrates a fundamental tradeoff in the SSL task. A low masking ratio creates a pretext task that is too simple, providing an insufficient supervisory signal for the model to learn meaningful cross-view correlations. Conversely, an excessively high masking ratio makes the reconstruction task too difficult, as there is not enough visible information to reliably infer the content of the masked views, which can hinder model convergence. Thus, an optimal ratio, such as 0.4 for AMiner, strikes a balance, creating a task that is challenging enough to compel the model to learn robust and expressive representations, yet solvable enough to ensure stable training.

\subsubsection{Impact of the Length of Meta-Paths}

\begin{figure}[t]
  \centering
  \includegraphics[width=0.87\columnwidth]{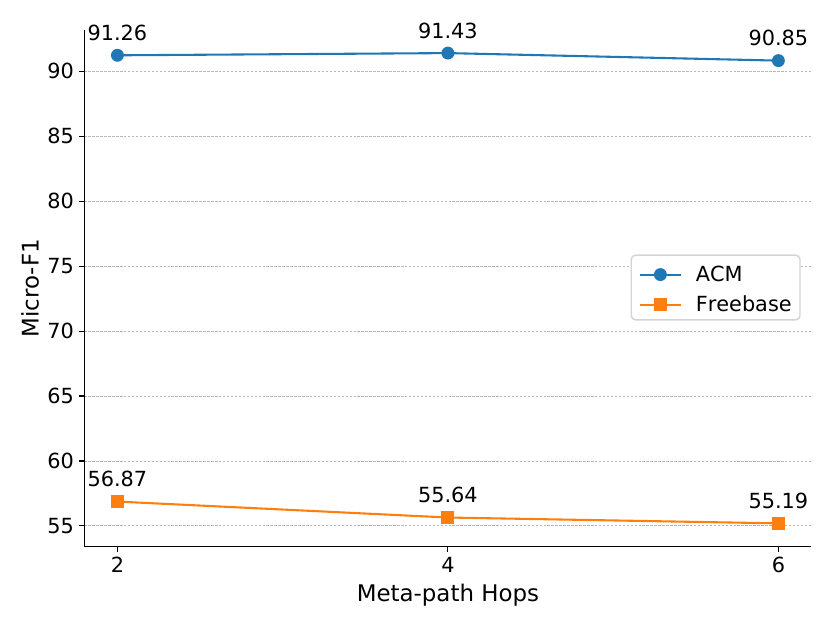}
    \caption{Sensitivity analysis of the hop of meta-paths on the ACM and Freebase datasets.}
  \label{fig:metapath_hop}
    \end{figure}

We further investigate the influence of meta-path length, or hop count, on our model's performance. For instance, the meta-path P-A-P has a length of 3 and a hop count of 2.
The length of a meta-path is a critical factor, as there is an inherent tradeoff---shorter paths may capture cleaner but limited neighborhood information, whereas longer paths can encapsulate richer, higher order semantics but risk introducing noise.
To explore this, we conduct experiments on the ACM and Freebase datasets using meta-paths of varying lengths. In particular, for the ACM dataset, we employ 2-hop \{P-A-P and P-S-P\}, 4-hop \{P-A-P-A-P and P-S-P-S-P\}, and 6-hop \{P-A-P-A-P-A-P and P-S-P-S-P-S-P\} sets. For the Freebase dataset, the corresponding sets are \{M-A-M and M-D-M\}, \{M-A-M-A-M and M-D-M-D-M\}, and \{M-A-M-A-M-A-M and M-D-M-D-M-D-M\}. The results, as illustrated in Fig. \ref{fig:metapath_hop}, reveal dataset-specific trends.
On Freebase, performance consistently degrades as the meta-path length increases, suggesting that the most direct 2-hop relationships already provide sufficient semantic information and longer paths primarily introduce noise.
Conversely, on the ACM dataset, performance peaks at 4-hop before declining. This suggests that 2-hop paths are insufficient to capture the most effective semantics for this domain, while 6-hop paths begin to accumulate irrelevant information that hinders model convergence. These findings confirm that the semantic payload of a meta-path view is heavily dependent on its length, a crucial consideration for downstream tasks.

\section{Future Work}

{
While MVMAE provides an efficient and effective framework for heterogeneous graph masking, several promising directions remain for future exploration. First, we plan to investigate hybrid training objectives by integrating view-level reconstruction with node-level contrastive alignment. This could potentially enhance representation robustness in domains where meta-path semantics are inherently weak or noisy. Second, to address the limitations of nondifferentiated masking, we aim to explore learnable masking strategies (e.g., via Gumbel--Softmax or reinforcement learning) that allow the model to adaptively select informative components to mask. Finally, extending our current paradigm to jointly incorporate structural and feature masking across multiple views may capture complementary information more effectively. While these enhancements may introduce additional computational overhead, finding an optimal balance between model complexity and performance gain will be a key focus of our subsequent research.
}

\section{Conclusion}

In this work, we propose MVMAE, a novel self-supervised framework for heterogeneous graphs designed to overcome the critical efficiency and effectiveness limitations of prior MAE methods. By shifting the learning paradigm from low-level feature masking to a high-level, multiview mask-and-reconstruct strategy, MVMAE learns robust, semantically rich node representations. Our design incorporates an efficient offline aggregation mechanism that eliminates the primary computational bottleneck of subgraph construction, enabling remarkable scalability.
Experimental results demonstrate that MVMAE establishes a new state-of-the-art, outperforming strong self-supervised and semisupervised baselines across six datasets.

{\appendix[Hyperparameters]
{To facilitate the reproduction of our work, we provide the hyperparameter settings for MVMAE across various datasets, as presented in Table~\ref{tab:hyperparameter}. The metapath set ($\Phi$) we use is presented in Table~\ref{tab:statistics}.}

\begin{table}[ht]
\caption{Hyperparameter settings of MVMAE on various datasets.}
\label{tab:hyperparameter}
\centering
\resizebox{\columnwidth}{!}{
\begin{tabular}{ccccccc}
\toprule
Dataset & ACM & DBLP & Freebase & AMiner & OAG & RCDD \\
\midrule
Masking ratio ($\rho$) & 0.2 & 0.3 & 0.5 & 0.5 & 0.3 & 0.5 \\
Embedding dimension ($d$) & 256 & 256 & 256 & 256 & 128 & 128 \\
Encoder depth ($L_e$) & 3 & 1 & 2 & 3 & 2 & 2 \\
Decoder depth ($L_d$) & 1 & 1 & 1 & 1 & 1 & 1 \\
Learning rate ($\eta$) & 1e-3 & 1e-3 & 1e-3 & 1e-3 & 1e-3 & 1e-3 \\
Dropout rate & 0.4 & 0.5 & 0.4 & 0.2 & 0.2 & 0.2 \\
\bottomrule
\end{tabular}
}
\end{table}
}

{\appendices
\section*{Comparisons to some related works}

{
In this section, we compare our multiview mask-then-reconstruct strategy to the strategies used in Hetero2Net \cite{li2025heterophily} and MV-HetGNN \cite{DBLP:journals/tkde/ShaoXWWZZ23}.

\subsubsection{Comparison with Hetero2Net \cite{li2025heterophily}}
While both MVMAE and Hetero2Net utilize a masking strategy, they fundamentally differ in their scope, objectives, and computational paradigms.
\subsubsection{Microlevel Versus Macrolevel Masking} Hetero2Net's ``masked meta-path prediction'' typically masks specific edges within a selected meta-path context to reconstruct local signals. On the contrary, MVMAE adopts a macrolevel strategy: it treats each meta-path-induced subgraph as an independent view and randomly masks the entire meta-path view.
\subsubsection{Low-Order Structure Versus High-Order Semantics} Hetero2Net focuses on a low-order structural  reconstruction task, while MVMAE focuses on semantic alignment. By forcing the model to reconstruct the high-level semantic representation of a fully masked view using the remaining visible views, MVMAE shifts the focus from local connections to inter-view semantic correlations.
\subsubsection{Advantages of our MVMAE Strategy}
Stronger High-order Semantic Capture: The cross-view consistency modeling forces the model to learn complex intrinsic correlations among different meta-paths, capturing deeper heterogeneous semantics.
\paragraph{Higher Computational Efficiency} Hetero2Net relies on complex subgraph sampling to control the neighborhood explosion caused by the ``small-world effect,'' which inevitably leads to structural information loss. MVMAE avoids this by using offline aggregation to precompute view representations prior to the view-level masking, significantly reducing computational overhead and scaling to large heterogeneous graphs.
\paragraph{More Effective Supervision} The full-view masking creates a more challenging self-supervised pretext task, compelling the encoder to extract highly discriminative features for downstream tasks.

\subsubsection{Comparison with MV-HetGNN \cite{DBLP:journals/tkde/ShaoXWWZZ23}}
While MVMAE and MV-HetGNN share a similar motivation of utilizing meta-paths for message passing, there are two crucial distinctions.
\paragraph{Decoupled Offline Processing Versus Coupled Online Training} MV-HetGNN performs message passing online during training by passing learnable node representations via meta-paths. Our MVMAE, however, performs message passing offline as a preprocessing step on static initial node features. This architectural difference is critical for scalability.
\paragraph{View-level Masking Versus Standard Autoencoding} MV-HetGNN uses a standard autoencoder to enhance intermediate representations without any masking strategy. MVMAE creatively employs a meta-path view masking approach. Reconstructing the missing views from visible ones explicitly drives the model to capture inter-view semantic dependencies, leading to more robust representations than standard autoencoding.
}
}

\clearpage
\raggedbottom
\makeatletter
\def\@textbottom{\vskip\z@ plus 1fill}
\makeatother

\begin{IEEEbiography}[{\includegraphics[width=1in,height=1.25in,keepaspectratio]{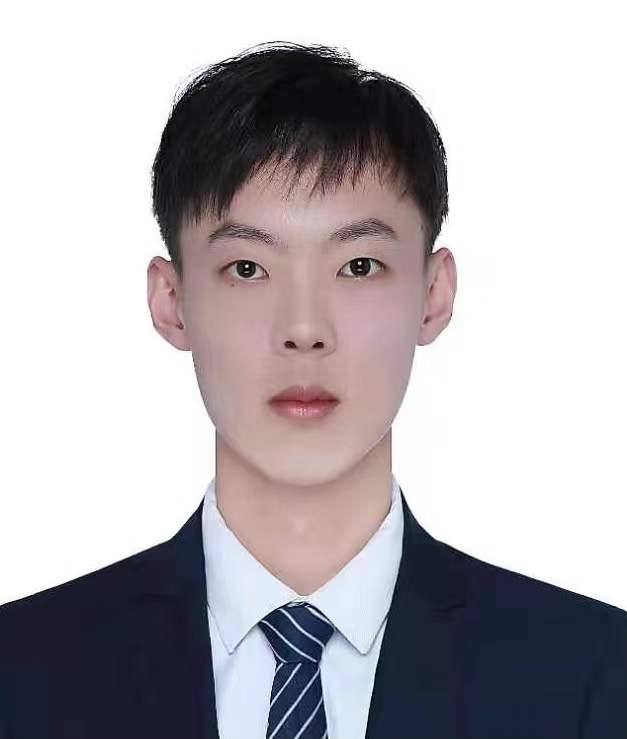}}]{Hao Geng}
received the B.S. degree in computer science and engineering from Soochow University, Suzhou, China, in 2021. He is currently pursuing the fourth-year Ph.D. degree with the School of Computer Science and Engineering with Beihang University, Beijing, China. His research interests include graph neural networks, large language models, and recommendation systems.
\end{IEEEbiography}

\begin{IEEEbiography}[{\includegraphics[width=1in,height=1.25in,keepaspectratio]{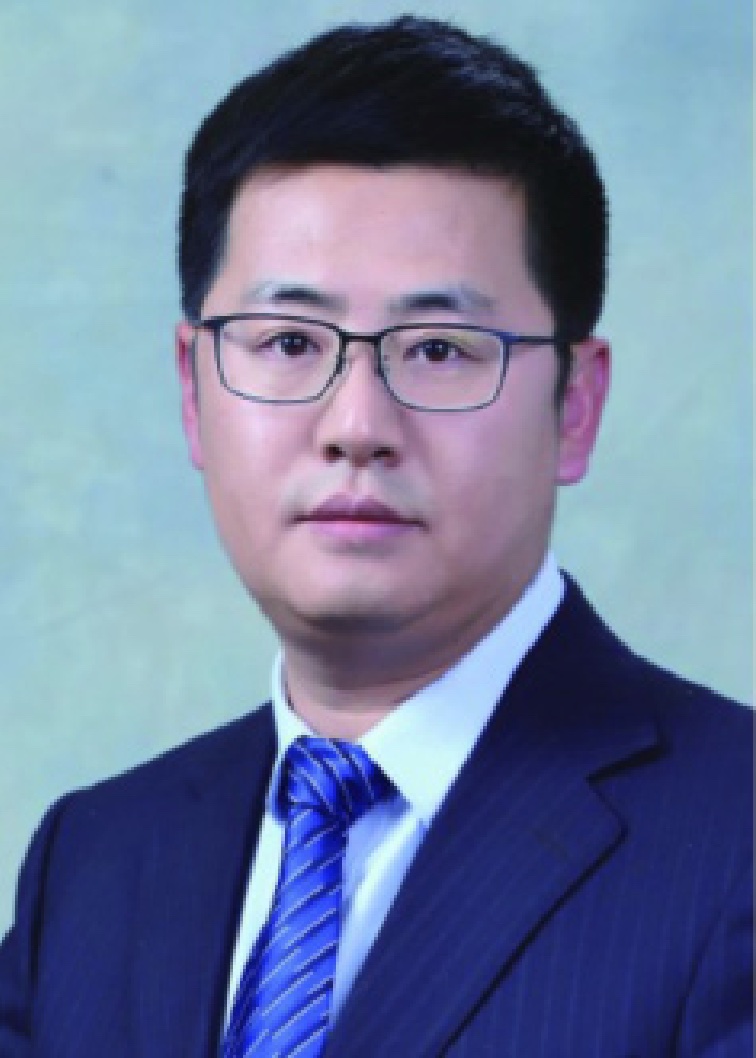}}]{Deqing Wang}
received the Ph.D. degree in computer science from Beihang University, Beijing, China, in 2013.
He is currently an Assistant Professor with the School of Computer Science, Beihang University, where he is the Deputy Chief Engineer with the National Engineering Research Center for Science Technology Resources Sharing and Service. He is also with the Zhongguancun Laboratory, Beijing. His current research interests include text categorization and data mining for software engineering and machine learning.
\end{IEEEbiography}

\begin{IEEEbiography}[{\includegraphics[width=1in,height=1.25in,keepaspectratio]{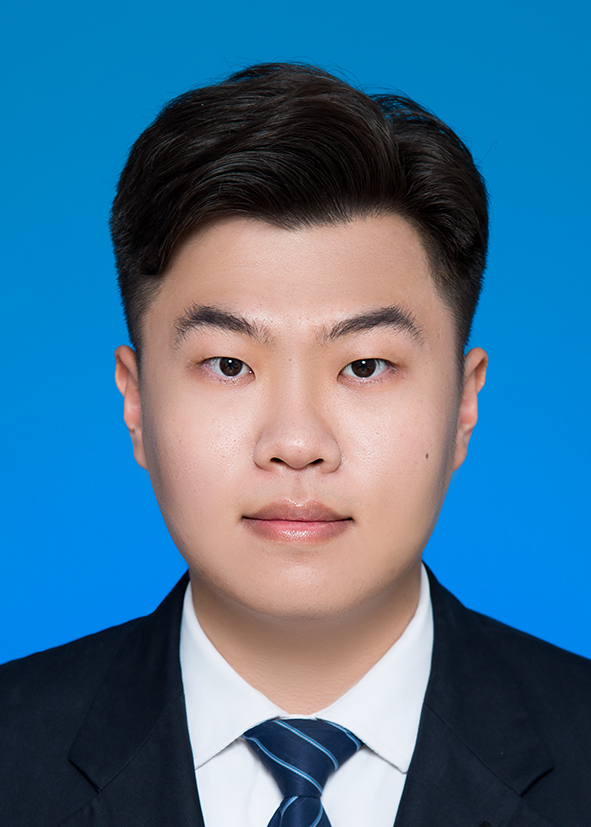}}]{Lihao Ge}
received the master's degree in computer science from Beihang University, Beijing, China, in 2025. He is currently with Tencent, Beijing, China. His research interests include graph neural networks and machine learning.
\end{IEEEbiography}

\begin{IEEEbiography}[{\includegraphics[width=1in,height=1.25in,keepaspectratio]{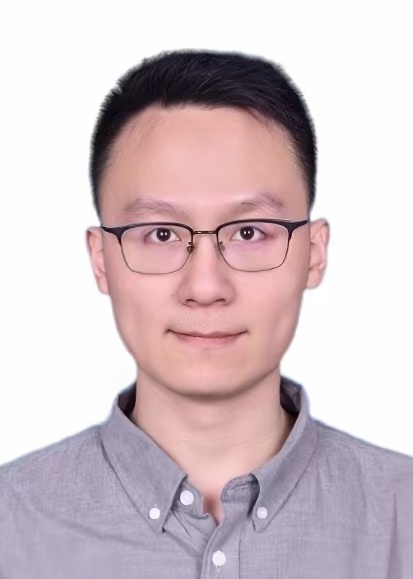}}]{Chenguang Du}
received the Ph.D. degree in computer science from Beihang University, Beijing, China, in 2024. He is currently an Assistant Researcher with the Zhongguancun Laboratory, Beijing. His research interests include graph neural networks and AI for networks.
\end{IEEEbiography}

\begin{IEEEbiography}[{\includegraphics[width=1in,height=1.25in,keepaspectratio]{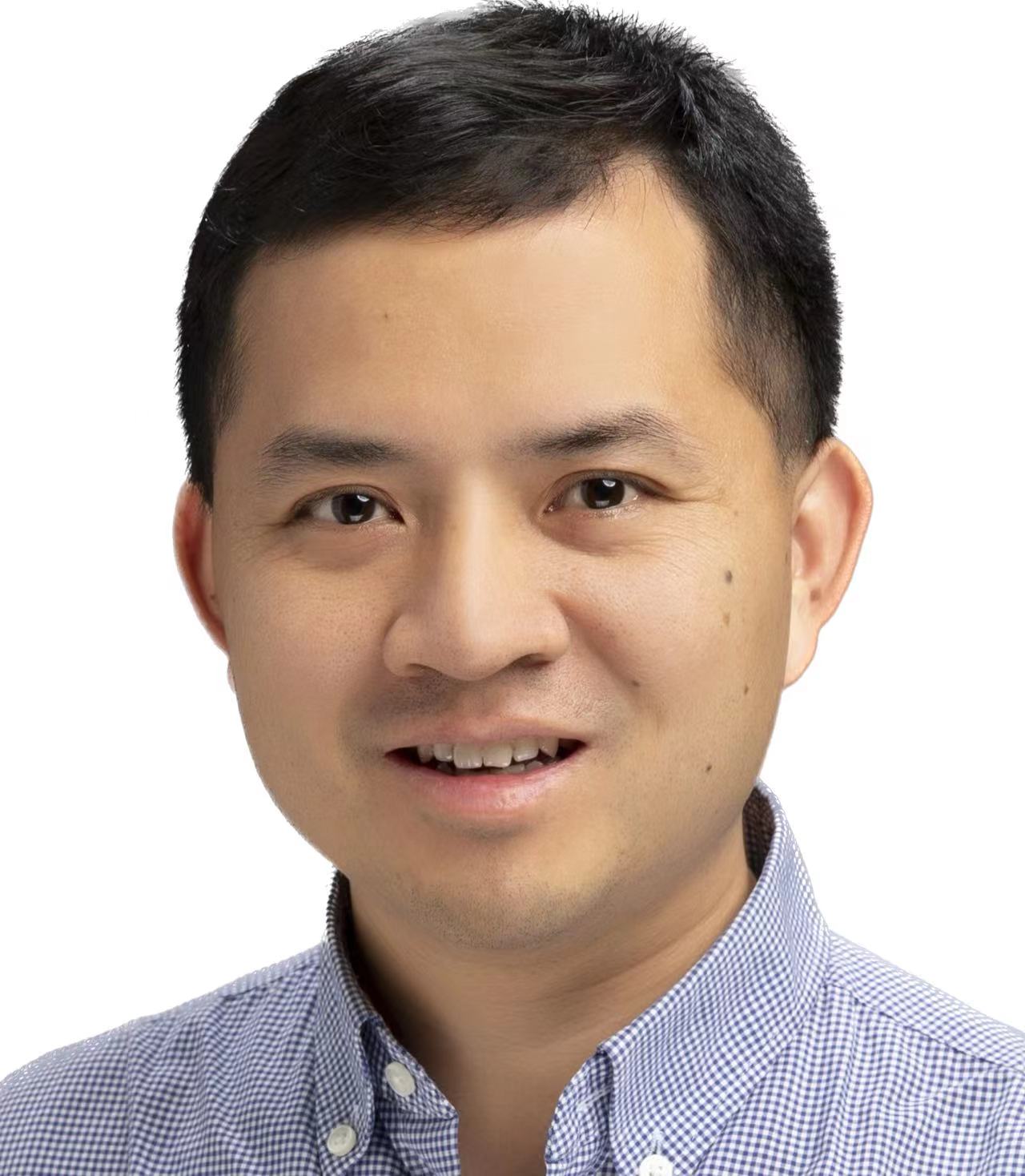}}]{Fuzhen Zhuang}
(Senior Member, IEEE) received the B.E. degree from the College of Computer Science, Chongqing University, Chongqing, China, in 2006, and the Ph.D. degree in computer science from the Institute of Computing Technology, Chinese Academy of Sciences, Beijing, China, in 2011. He is currently a Full Professor with the Institute of Artificial Intelligence, Beihang University, Beijing. He is also with the Zhongguancun Laboratory, Beijing. He has published more than 140 articles in some prestigious refereed journals and conference proceedings such as the Nature Communications, IEEE Transactions on Knowledge and Data Engineering, IEEE Transactions on Cybernetics, IEEE Transactions on Neural Networks and Learning Systems, ACM Transactions on Knowledge Discovery from Data, ACM Transactions on Intelligent Systems and Technology, Information Sciences, Neural Networks, SIGKDD, IJCAI, AAAI, TheWebConf, SIGIR, ICDE, ACM CIKM, ACM WSDM, SIAM SDM, and IEEE ICDM. His research interests include transfer learning, machine learning, data mining, multitask learning, knowledge graph, and recommendation systems. Dr. Zhuang is a Senior Member of the CCF. He was a recipient of the Distinguished Dissertation Award of CAAI in 2013.
\end{IEEEbiography}

\begin{IEEEbiography}[{\includegraphics[width=1in,height=1.25in,keepaspectratio]{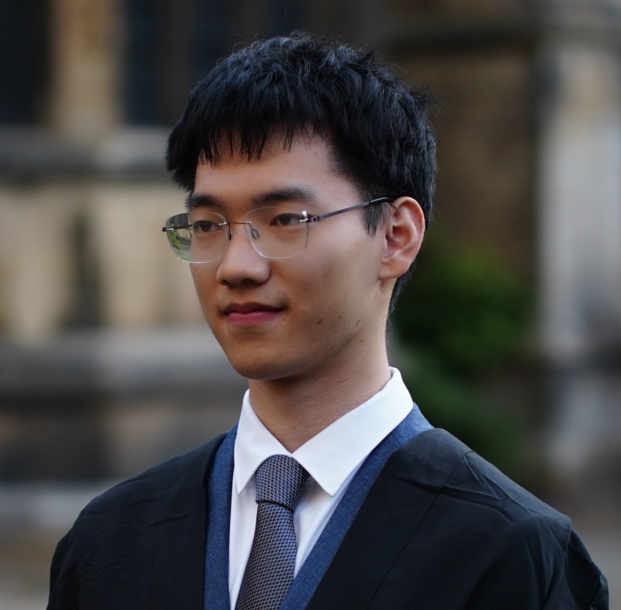}}]{Bowen Yang}
received the bachelor's degree in mathematics from Beihang University (BUAA), Beijing, China, in June 2025, and the M.Phil. degree in scientific computing from the University of Cambridge, Cambridge, U.K., in August 2026. He is currently working as a Research Assistant with Westlake University, Hangzhou, China.
\end{IEEEbiography}

\begin{IEEEbiography}[{\includegraphics[width=1in,height=1.25in,keepaspectratio]{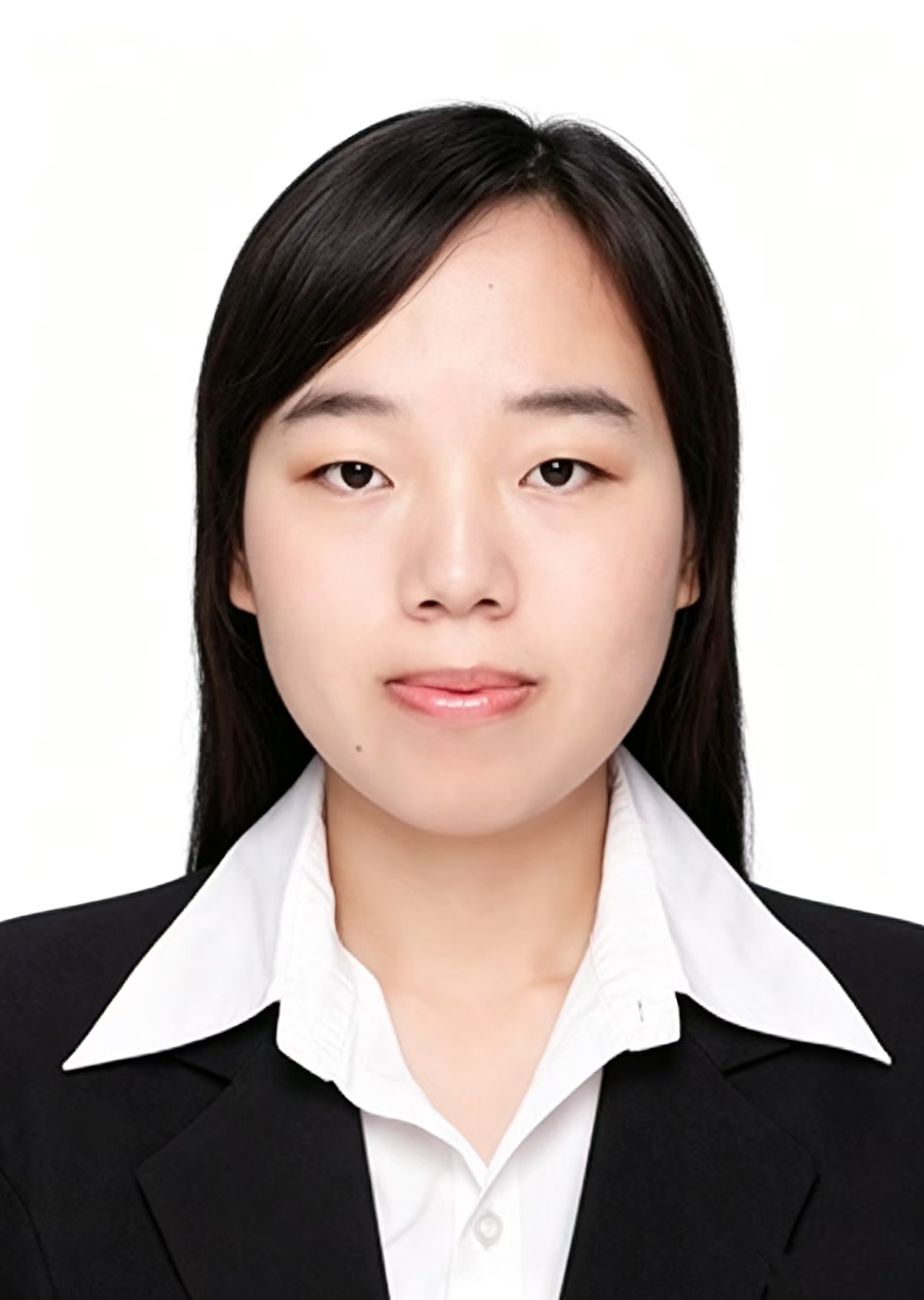}}]{Yanlin Hu}
received the master's degree in computer science from Beijing Institute of Technology, Beijing, China, in 2016. She is currently with the National Computer Network Emergency Response Technical Team/Coordination Center of China (CNCERT/CC), Beijing. Her research interests include natural language processing and high volume language information processing.
\end{IEEEbiography}

\begin{IEEEbiography}[{\includegraphics[width=1in,height=1.25in,keepaspectratio]{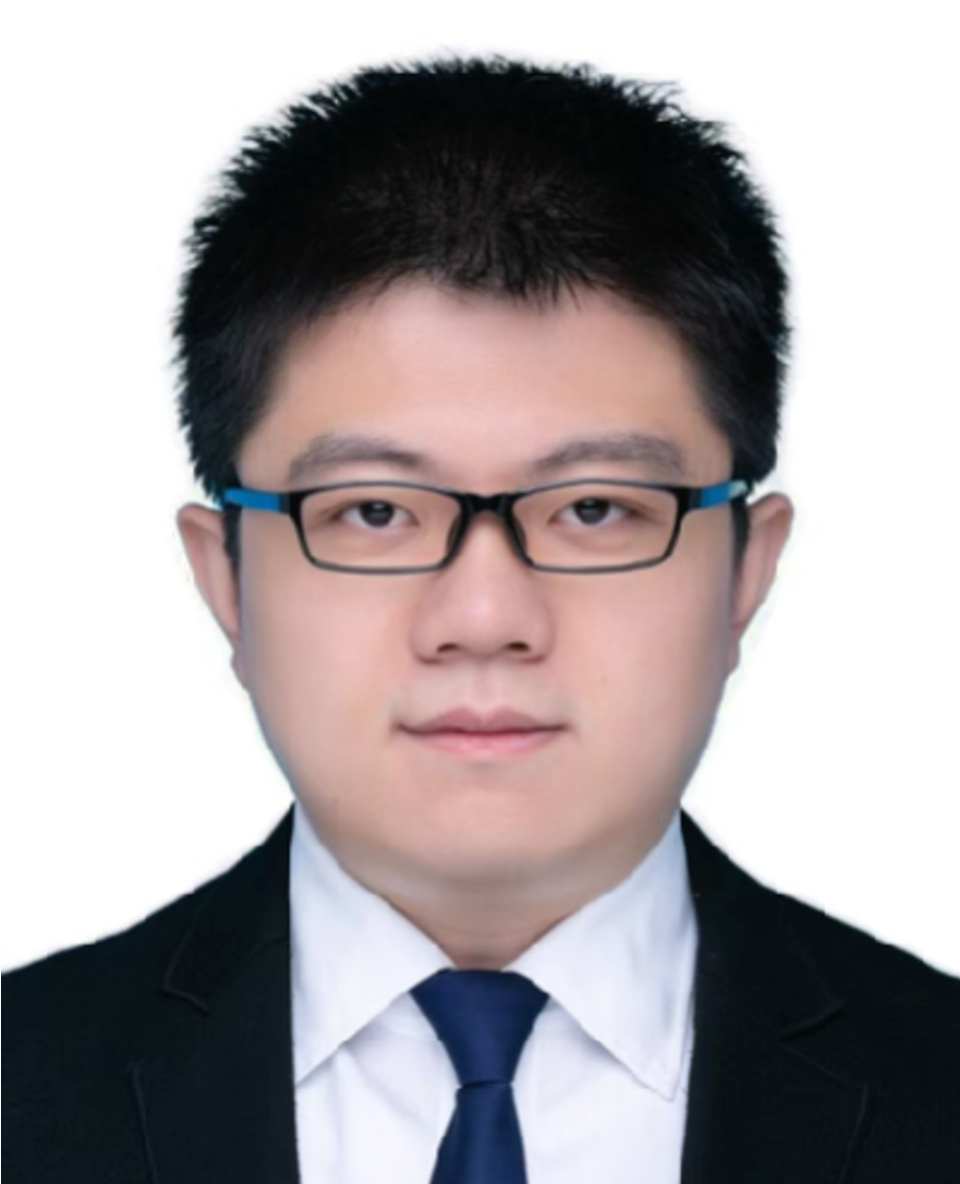}}]{Songtao Wang}
received the master's degree in systems on chip from the University of Southampton, Southampton, U.K., in December 2011, and the Ph.D. degree in computer science and technology from Tsinghua University, Beijing, China, in July 2020. He is currently an Associate Researcher with Zhongguancun Laboratory, Beijing, China. His research interests include distributed machine learning and source address validation for Internet networks.
\end{IEEEbiography}

\end{document}